\documentclass[11pt]{article}
\usepackage{fullpage}
\usepackage{algorithm}
\usepackage{geometry}

\usepackage{microtype}
\usepackage{graphicx}
\usepackage{subfigure}
\usepackage{booktabs} % for professional tables
\usepackage{algpseudocode}
\usepackage{booktabs}  % For \toprule, \midrule, \bottomrule
\usepackage{makecell}

\usepackage{hyperref}
\usepackage{natbib}
\usepackage[utf8]{inputenc} % allow utf-8 input
\usepackage[T1]{fontenc}    % use 8-bit T1 fonts
\usepackage{hyperref}       % hyperlinks
\usepackage{url}            % simple URL typesetting
\usepackage{amsfonts}       % blackboard math symbols
\usepackage{nicefrac}       % compact symbols for 1/2, etc.
\usepackage{microtype}      % microtypography
\usepackage{xcolor}         % colors
\usepackage{enumitem}
\usepackage[toc,page,header]{appendix}
\usepackage{minitoc}

\algnewcommand{\LeftComment}[1]{\(\triangleright\) {\color{LightCyan}{#1}}}

\usepackage{amsmath}
\usepackage{amssymb}
\usepackage{mathtools}
\usepackage{amsthm}
\usepackage{pifont}
\usepackage{etoc}
\etocdepthtag.toc{mtchapter}
\etocsettagdepth{mtchapter}{subsection}
\etocsettagdepth{mtappendix}{none}
\usepackage[capitalize,noabbrev]{cleveref}

\makeatletter
\newtheorem*{rep@theorem}{\rep@title}
\newcommand{\newreptheorem}[2]{%
\newenvironment{rep#1}[1]{%
 \def\rep@title{#2 \ref{##1}}%
 \begin{rep@theorem}}%
 {\end{rep@theorem}}}
\makeatother

\definecolor{royalblue}{rgb}{0.25, 0.41, 0.88}
\hypersetup{
  colorlinks,
  citecolor=royalblue,
  linkcolor=royalblue,
  urlcolor=royalblue}  

\theoremstyle{plain}
\newtheorem{theorem}{Theorem}[section]
\newtheorem{proposition}[theorem]{Proposition}

\newtheorem{corollary}[theorem]{Corollary}

\newtheorem{example}[theorem]{Example}
\newtheorem{remark}[theorem]{Remark}

\newreptheorem{theorem}{Theorem}
\newreptheorem{lemma}{Lemma}
\newreptheorem{proposition}{Proposition}
\newreptheorem{corollary}{Corollary}

\usepackage{pgfplotstable}
 \pgfplotsset{compat=1.18}
\usepackage{pifont}

\usepackage{tcolorbox}

\usepackage{xspace}
\usepackage[colorinlistoftodos, color=blue!30!white %   ,shadow
]{todonotes}                                        
\newcommand{\leoni}[2][]{\todo[size=tiny,color=green!20!white,#1]{LW: #2}\xspace}

\newcommand{\N}{\ensuremath{\mathbb{N}}\xspace}
\newcommand{\Z}{\ensuremath{\mathbb{Z}}\xspace}

\newcommand{\R}{\ensuremath{\mathbb{R}}\xspace}

\newcommand{\MM}{\mathcal{M}}

\newcommand{\Prob}{\ensuremath{\mathbb{P}}\xspace}
\newcommand{\E}{\ensuremath{\mathbb{E}}\xspace}
\newcommand{\1}{1\hspace{-0,9ex}1}

\newcommand*\diff{\mathop{}\!\mathrm{d}}

\newcommand{\Tau}{\mathrm{T}}
\newcommand{\Eta}{\mathrm{H}}

\newcommand{\Leb}{\mathrm{Leb}}
\newcommand{\hnu}{\hat{\nu}}

\newcommand{\G}{\mathbb{G}}

\newcommand{\mcf}{\mathcal{F}}

\newcommand{\mcu}{\mathcal{U}}
\newcommand{\mcx}{\mathcal{X}}
\newcommand{\mcy}{\mathcal{Y}}
\newcommand{\mcz}{\mathcal{Z}}

\newcommand{\mvert}{\,|\,}
\newcommand{\abs}[1]{|#1|}
\newcommand{\norm}[1]{\lVert#1\rVert}

\xdefinecolor{dpurple}{rgb}{0.6,0,0.6}
\title{Private Synthetic Graph Generation and Fused Gromov-Wasserstein Distance}

\author{%
 Leoni Carla Wirth$^{1}$
 \and
 Gholamali Aminian$^{2}$
 \and
 Gesine Reinert$^{2,3}$
}
\usepackage{times}
\begin{document}

\maketitle
\renewcommand\thefootnote{$^1$}\footnotetext{University of G\"ottingen, Institute for Mathematical Stochastics.}
\renewcommand\thefootnote{$^2$}\footnotetext{The Alan Turing Institute.}
\renewcommand\thefootnote{$^3$}\footnotetext{University of Oxford, Department of Statistics. }

\begin{abstract}
Networks are popular for representing complex data. In particular, differentially private synthetic networks are much in demand for method and algorithm development. The network generator should be easy to implement and should come with theoretical guarantees. Here we start with complex data as input and jointly provide a network representation as well as a synthetic network generator. Using a random connection model, 
we  devise  an effective algorithmic approach for generating attributed synthetic graphs  which is $\epsilon$-differentially private at the vertex level, while preserving  utility under an appropriate notion of distance which we develop. 
We provide theoretical guarantees for the  accuracy of the private synthetic graphs using the  fused Gromov-Wasserstein distance, which extends the Wasserstein metric to structured data. Our method draws inspiration from the  PSMM method of \citet{he2023}.
%Our method runs in .... time .... and demonstrates improved accuracy compared to prior approaches.
\end{abstract}
\tableofcontents
% \begin{keywords}%
%  Differential privacy, synthetic network generation, 
%  {networks with attributes}, Stein's  method
% \end{keywords}

\section{Introduction}

Networks are a popular means for representing  complex data, see for example \citet{rathkopf2018network}.  Synthetic networks can be used to simulate particular behavior, and also to assess whether the network contains unusual features, or anomalies. Synthetic networks are furthermore applied to augment data when the underlying data set is highly imbalanced. They are also used for method and algorithm development.  
A multitude of synthetic network generators are available, including parametric methods, as in  \cite{batagelj2005efficient}, empirical approaches such as  \citet{reinert2024steingen}, and deep learning approaches, see for example \cite{qian2023synthcity}. 

Synthetic networks are also used for sharing data sets which contain sensitive features, which is often the case for example in networks of financial transaction or in health-related data,  to share data which are semantically and statistically similar to the real data, while protecting privacy. However, synthetic data are not automatically privacy-perserving, see for example \citet{houssiau2022tapas}. A notion of privacy which is often used in this context is {\it differential privacy} from  \cite{dwork2006differential}; it is a property of a randomized data generating mechanism (not of a data set) and, at high level, ensures that when two input data sets differ slightly, then the distributions of the synthetically generated data based on each of the two input data sets are also close. 

To achieve differential privacy, many strategies have been suggested, but obtaining theoretical guarantees for these methods remains a challenge, see \cite{zhang2025privdpr} for an overview. 
In \citet{he2023} 
a synthetic data generation algorithm called ${\rm PSMM}$ is developed for generating differentially private synthetic data in a bounded metric space by perturbing the feature space and adding random noise to obtain a possibly signed measure;
then the algorithm finds a probability measure that minimised the bounded Lipschitz distance to the
perturbed measure. 
In \cite{he2023} it is proven that this algorithm  has near-optimal utility guarantees.  

A key issue for differential privacy is how to assess closeness between distributions. The definition of differential privacy by \cite{dwork2006differential} assesses the difference between distributions over all possible sets of outcomes (the {\it total variation distance}). 
In many instances, this notion is too strong. For example in a spatial network, two networks for which the spatial coordinates differ by only a tiny amount would be judged to be just as far away as two networks which are set in different parts of the underlying space. Instead, \cite{he2023} measure the utility of the output by the expected 1-Wasserstein distance between the generated distributions. 

The input for synthetic network generators is usually taken to be an observed network. However the observed network is itself only a representation of a complex data set. A key novelty of our approach is that we directly take the complex data as input and jointly generate a ``true'' network and differentially private synthetic networks. 

For achieving differential privacy, 
our paper builds on \citet{he2023} and extends it in the following directions. 
First, \citet{he2023} use  the bounded Lipschitz metric for the synthetic data generation. %This metric is not always easy to compute. Instead we 1-Wasserstein distance assesses the distance between distributions in terms of their behaviour on Lipschitz-continuous functions. Lipschitz continuity is a natural notions for data in Euclidean space. However when the underlying space in non-Euclidean, such as in the setting of networks, other notions of distance are required. Synthetic networks which are generated for privacy-preserving purposes in which the vertices often do have attributes, which may even be the main focus of interest. Hence, in this paper we introduce a new metric between synthetic generators of networks with vertex attributes. 
%It is a Wasserstein distance but the Lipschitz continuity is now  required not with respect to Euclidean distance, but instead with respect to 
 %a  fused Gromov-Wasserstein metric. This construction allows us to 
 Here we replace the often not easy to compute bounded Lipschitz distance between distributions by the easier to calculate total variation distance. We call the resulting algorithm for  differentially private synthetic data  generation  TV-PSMM. %\leoni{Do we call the network generator (Algorithm\ref{alg: graph generation}) or the adapted PSMM (Algorithm~\ref{alg: PSMM}) TV-PSMM?}

 Second, in \citet{he2023} only the expected 1-Wasserstein distance is considered. The 
 1-Wasserstein distance assesses the distance between distributions in terms of their behaviour on Lipschitz-continuous functions. Lipschitz continuity is a natural notions for data in Euclidean space. However when the underlying space in non-Euclidean, such as in the setting of networks, other notions of distance are required. Synthetic networks which are generated for privacy-preserving purposes in which the vertices often  have attributes, which may even be the main focus of 
of interest. Hence, in this paper we introduce a new metric between synthetic generators of networks with vertex attributes. 
It is a Wasserstein distance for which  the Lipschitz continuity is now  required not with respect to Euclidean distance, but instead with respect to 
 a  fused Gromov-Wasserstein metric. We are able to provide theoretical guarantees which assess the distributional distance, in our metric, between the distribution on networks generated by our differentially-private synthetic network generator, which we call PSGG, and the distribution which generated the original network data. Evaluating the expected distance is  a particular application of our more general result. 
Third, while \citet{he2023} employ discrete Laplacian perturbations, we consider more  general noise distributions. 

 The paper is structured as follows: Section \ref{sec:prem} introduces differerential privacy, fused Gromov-Wasserstein and total variation distances,  and the TV-PSMM algorithm. %$\epsilon$-differential private synthetic data algorithm.
 Then, we discuss our private synthetic graph generation (PSGG) algorithm in Section~\ref{sec:PSGG}. We derive our main theoretical results regarding the effect of the PSGG algorithm via two approaches and their rates in Section~\ref{sec:Main}. The conclusion and future works are presented in Section~\ref{sec:con}. Detailed proofs are deferred to the Appendix, where also more related work, and more illustrations, can be found.

\section{Preliminaries}\label{sec:prem}
\subsection{Differential Privacy}
Inspired by \cite{dwork2006differential}, we focus on $\epsilon$-differential privacy. A randomized algorithm $\MM$ provides \textit{$\epsilon$-differential privacy} if for any  inputs $A,A'$ that differ  on only one data point, it satisfies for any measurable set $S\in\mathrm{range}(\MM)$,
    \begin{equation}
        \frac{\mathbb{P}(\MM(A)\in S)}{\mathbb{P}(\MM(A')\in S)}\leq e^{\epsilon}.
    \end{equation}
\subsection{Fused Gromov-Wasserstein distance} \label{ssec: FGW}
The fused Gromov-Wasserstein (FGW) distance introduced in \cite{vayer2020} measures differences between structured objects with features. It combines a Gromov-Wasserstein distance for the %(graph) 
structural comparison with a Wasserstein metric for the feature comparison and can be seen as generalization of both metrics.  
The underlying idea of the fused Gromov Wasserstein distance is to represent the structured object with features as a measure $\mu = \sum_{i=1}^n h_i \delta_{(x_i,a_i)}$, where $(h_i)_{i\in[n]}\subset \R_+$ such that $\sum_{i=1}^n h_i = 1$. Here the $a_i\in\Omega$ encode the feature information while the structural information is captured by the $x_i$, which are assumed to %More precisely, the $x_i$ 
lie in a Polish metric space $(X,d_X)$ that represents the object structure.
%using the metric measure space approach.

We focus on the particular discrete case where the structured objects (with features) are given by attributed graphs, see \citet[Section~4]{vayer2020}. Then for two structured objects $\mu = \sum_{i=1}^n h_i \delta_{(x_i,a_i)}$ and $\eta = \sum_{j=1}^m g_j \delta_{(y_j,b_j)} $ the fused Gromov-Wasserstein (FGW) distance (or metric) with parameters $p,q\geq 1$ and $\alpha\in[0,1]$ is defined by 
\begin{align}\label{def:fgw}
    d_{FGW}(\mu,\eta)=d_{FGW,\alpha,p,q}(\mu,\eta) = \bigg(\min_{\pi\in\Pi(\mu,\eta)} E_{\alpha,p,q}(\pi)\bigg)^{\frac{1}{q}},
\end{align}
where $ E_{\alpha,p,q}(\pi):=\sum_{i,j,k,l} \big((1-\alpha) d(a_i,b_j)^q
    \alpha \abs{d_X(x_i,x_k) - d_Y(y_j,y_l)}^q\big)^p \pi_{k,l}\pi_{i,j}$
% \begin{align*}
%     %\big(\underbrace{(1-\alpha) d(a_i,b_j)^q}_{\text{WS comparison attributes}} + %\underbrace{\alpha \abs{d_X(x_i,x_k) - d_Y(y_j,y_l)}^q\big)^p }_{\text{Gromov WS comparison edge structure}}  %\underbrace{\pi((x_k,a_k),(y_l,b_l))}_{\text{mass transported from $a_k$ to $b_l$}}
% \end{align*}
for $\pi_{i,j} = \pi((x_i,a_i),(y_j,b_j))$,
and $\Pi(\mu,\eta) := \{\pi\in \R_+^{n\times m}\mvert \sum_i \pi_{ij} = b_j; \sum_j \pi_{ij} = a_i\}$ is the set of {\it couplings}; a coupling between two probability measures is  a suitable construction of the measures on a common probability
space. Note that the first equality in \eqref{def:fgw} can be interpreted as Wasserstein comparison of the attributes with respect to an underlying metric $d = d_{\Omega}$, while the equality term can be viewed as a Gromov-Wasserstein comparison of the edge structure. Furthermore, $\pi_{i,j}$ gives the amount of mass that is transported from $i$ to $j$ while $\alpha$ can be seen as trade-off parameter between attribute and structure cost. We refer to \cite{vayer2020} for a thorough introduction of the FGW distance.

The FGW distance can also be used to measure the distance between random attributed graph distributions. More precisely, we can consider the 1-Wasserstein metric with respect to the FGW distance to obtain a measure of distance between any two graph distributions $P^{\text{M}}$ and $P^{\Eta}$. By duality of the 1-Wasserstein distance this is equivalent to defining 
\begin{equation*} %\label{eq: integral metric}
    d_{\mcf}(P^{\text{M}},P^{\Eta}) := \sup\limits_{f\in\mcf} \big\vert \E(f(\text{M})) - \E(f(\Eta))\big\vert,
\end{equation*}
for $\mcf$ the set of test functions $f:\G \to \R$ that are Lipschitz with respect to the FGW distance. This class of test functions, in particular contains the comparison to a reference graph $\xi$, i.e. the functions $f_{\xi}(\mu) = d_{FGW}(\xi,\mu)$ are Lipschitz with respect to the FGW distance. This is a direct consequence of the reverse triangle inequality.

For the rest of the paper we set $p=q=1$ to simplify the notation. However, our results hold for general $p,q\geq 1$. Furthermore, we assume that the cost of a single edge can be bounded by some $C>0$, i.e. $\abs{d_X(x_i,x_k) - d_Y(y_j,y_l)}\leq C$ for any $i,j,k,l$.

\subsection{Total variation distance} \label{ssec: TV}
The total variation norm of a signed measure~$\eta$ on some measurable space~$\mcz$ can be defined using the Jordan-Hahn decomposition of the signed measure $\eta = \eta^+ - \eta^-$ into the difference of two   positive measures $\eta^+$ and   $\eta^-$; %. More precisely,
we set
$$\norm{\eta}_{TV} := \abs{\eta}(\mcz) := \eta^+(\mcz) + \eta^-(\mcz).$$
In the special case where the signed measure~$\eta$ has a discrete support $\mcz = \{z_1,\ldots,z_n\}$, we have for every $Z\subset \mcz$ that
$\eta(Z) = \sum_{z\in Z} \eta(\{z\})$. This yields
$$\norm{\eta}_{TV} = %\eta^+(\mcz) + \eta^-(\mcz) = 
\sum_{z\in Z} \abs{\eta(\{z\})}.$$
In particular, the total variation norm induces the total variation distance $d_{TV}$ that for two signed measures $\eta_1$, $\eta_2$ on  $\mcz = \{z_1,\ldots,z_n\}$ is given by the easy to evaluate formula 
$$d_{TV}(\eta_1,\eta_2) := \norm{\eta_1-\eta_2}_{TV} = \sum_{z\in Z} \abs{\eta_1(\{z\})-\eta_2(\{z\})}.$$
%\vspace{-0.3in}
\subsection{An $\epsilon$- differential private synthetic data algorithm}
\label{ssec: Algorithm 1}
We use the total variation private signed measure mechanism (TV-PSMM), which is based on the private signed measure meachanism (PSMM) of \citet{he2023}, to generate differentially private synthetic data.  All proof details are deferred to App.~\ref{app:sec_prem}\,.

We briefly recall the PSMM algorithm proposed by \citet{he2023}. Given the true data $\mcx\subset \Omega$ and a partition $(\Omega_1,\ldots,\Omega_m)$ of $\Omega$, the PSMM first generates a new dataset $\mcy$ by sampling a representative $y_i$ in each cell~$\Omega_i$. By adding i.i.d.\,discrete Laplacian noise~$\lambda_1,\ldots,\lambda_m$ to the counts of true data points~$n_i$ in each cell~$\Omega_i$, a signed measure~$\nu$ that is supported on $\mcy$ is obtained. Finally, the private synthetic dataset is created by sampling $m'$  realizations with respect to the probability measure $\hnu$ that is the closest probability measure to $\mu$ with respect to the bounded Lipschitz distance. Note that the bounded Lipschitz distance can be seen as a generalization of the $1$-Wasserstein distance to signed measures, see \citet[Section~2]{he2023}.%\gr{add what the bL distance is?}

We modify the algorithm of \citet{he2023} by allowing more general noise, i.e. $\lambda_1,\ldots,\lambda_m\overset{i.i.d.\,}{\sim}P_\Lambda$ for $P_{\Lambda}$ the distribution of some %\todoa{$\R$ or $\Z$}\leoni{our results hold for general $\R$-valued rv. Due to the interpretation via vertex counts $\Z$ might be more suitable.}$\R$-valued 
scalar random variable $\Lambda$. Additionally, we adapt their procedure in the linear programming step to our setting; instead of using  the bounded Wasserstein distance, which is often not easy to evaluate, %. More precisely, 
we choose the closest probability measure $\hnu$ with respect to the total variation distance. Furthermore, we output this optimal probability measure~$\hnu$ instead of a private synthetic dataset sampled from $\hnu$. For completeness, we state the TV-PSMM algorithm before we give a new linear program in Algorithm~\ref{alg: linear_programming} to find the optimal probability measure $\hnu$.

\begin{algorithm}[!ht]	
\caption{Total Variation Private Signed Measure Mechanism (TV-PSMM) } \label{alg: PSMM}
\begin{algorithmic}
\State {\textbf{Input:}}
 true data $\mcx=(x_1,\ldots,x_n) \in \Omega^n$,  partition $(\Omega_1,\dots,\Omega_m)$ of $\Omega$, privacy parameter $\epsilon >0$, noise distribution $P_{\Lambda}$.

\begin{description}
  \item[Compute the true counts] Compute the true count in each set of the partition, %regime 
  $n_i = \#\{x_j\in \Omega_i:j\in [n]\}$.
  \item[Create a new dataset] For each $i\in [n]$, uniformly choose an element $y_i\in\Omega_i$  {  independently of $\mathcal X$}, and let  $\mcy$ be the collection of $n_i$ copies of $y_i$.
  \item[Add noise]   Choose i.i.d.\, $\lambda_i\sim P_\Lambda$; 
  %i.i.d.\,. 
  perturb the empirical measure $\mu_{\mcy}$ of $\mcy$  to obtain a signed measure $\nu$ given by 
  %such that 
  \[\nu (\{y_i\}) := (n_i + \lambda_i)/n.\]
  \item[Linear programming] Using Algorithm~\ref{alg: linear_programming}, find the closest probability measure $\hat{\nu}$ of $\nu$ with respect to the total variation distance. %, and generate synthetic data $\widehat{\mcy}$ from $\hat{\nu}$.
\end{description}

\State {\textbf{Output:}} probability measure $\hnu$ on $\mcy$ describing the distribution of synthetic data. %synthetic data 
  %$\widehat{\mcy} = (y_1,\ldots,y_{m'}) \in \Omega^{m'}$ for some integer $m'$.
\end{algorithmic}
\end{algorithm}

\begin{algorithm}[!ht]			
\caption{Linear Programming} \label{alg: linear_programming}
\begin{algorithmic}
\State {\textbf{Input:}}  
 A discrete signed measure $\nu$ supported on $\mcy=\{y_1,\dots,y_m\}$.
\begin{description}
    \item[Solve the linear program] Solve the linear programming problem with $2m$ variables and $3m+1$ constraints:
    \begin{align*}
        \min \quad & \sum_{i=1}^m u_i\\
        \mbox{s.t.}\quad &u_i \geq \nu(\{y_i\}) - \tau_i,  \quad u_i \geq \tau_i-\nu(\{y_i\}), \quad \tau_i \geq 0,\quad\forall i\leq m;    \quad 
        %\\
         %&
         \sum_{i=1}^m \tau_i = 1.
    \end{align*}
\end{description}

\State {\textbf{Output:}} a probability measure $\hat{\nu}$ with $\hat\nu(\{y_i\})=\tau_i $.

\end{algorithmic}
\end{algorithm}
Following the proof of Proposition 5 in \citet{he2023}, we can derive the following result directly.
\begin{proposition}\label{prop:DP-gaurantee}
 If for all integers $k$ in the support of $P_\Lambda$ and for $a \in \{-1, 1\},$
 $$\frac{ P_\Lambda(k+ a )}{P_\Lambda(k)} \le e^\epsilon, 
 $$
 then 
 %$P_\Lambda=\mathr%{Lap}_{\mathbb{Z}}(1/\epsilon)$ is the discrete Laplace distribution, then 
 the Algorithm~\ref{alg: PSMM} is $\epsilon$-differential private.
\end{proposition}
%\gr{new result here, please have a look}\leoni{I checked and didn't found a mistake/problem. }
\begin{example}
If $P_\Lambda=\mathrm{Lap}_{\mathbb{Z}}(1/\epsilon)$ is the discrete Laplace distribution, then the assumption of Proposition~\ref{prop:DP-gaurantee} is satisfied. A second example in which the assumption is satisfied is that of 
 $P_\Lambda(k)  = |k|^\epsilon Z(A, \epsilon)$ for $1 \le  |k| \le A$, with  $Z(A,\epsilon) $ the normalising constant, and $A\in \N$.
\end{example}

We further prove that the output of Algorithm~\ref{alg: linear_programming} is the closest probability measure to the discrete measure in total variation distance.
% ...\todoa{pls explain here the main message of next proposition.}
\begin{proposition}\label{prop:total_measure}
Algorithm~\ref{alg: linear_programming} outputs the  probability measure $\hnu$ on $\mcy$ that is closest to the given discrete signed measure~$\nu$ on $\mcy$ with respect to the total variation distance.
\end{proposition}
% \todoa{Do we want to keep this proof here? is it different from \citep{he2023}?}\leoni{It is different from \cite{he2023} as we have the total variation distance. However, the argument is fairly standard. So from my point of view it is not necessary to give the proof here.}\todoa{Tnx.}
% \begin{proof}
% Let $\nu$ be a discrete signed probability measure on $\mcy$. Then for any probability measure $\tau$ on $\mcy$ the total variation distance between $\nu$ and $\tau$ is given by
% $$d_{TV}(\nu,\tau) = \sum_{i=1}^m \abs{\nu_i - \tau_i},$$
% where $\nu_i := \nu(\{y_i\})$ and $\tau_i := \tau(\{y_i\})$, see Section~\ref{ssec: TV}. Thus, finding the closest probability measure with respect to the total variation distance is equivalent to solving the program~\eqref{eq: P Algorithm2} given by
% \begin{align} \label{eq: P Algorithm2}
%         \min \quad & \sum_{i=1}^m \abs{\nu_i - \tau_i} \tag{P$1$}\\
%         \mbox{s.t.}\quad & \sum_{i=1}^m \tau_i = 1, \quad \tau_i \geq 0, \quad \forall i\leq m.
% \end{align}
% Then as $\abs{\nu_i - \tau_i} = \max\{\nu_i - \tau_i,  \tau_i -\nu_i\}$, minimizing the absolute value~$\abs{\nu_i - \tau_i}$ is equivalent to finding the minimal $u_i\geq \max\{\nu_i - \tau_i,  \tau_i -\nu_i\}$. In particular, \eqref{eq: P Algorithm2} is equivalent to solving the linear program in Algorithm~\ref{alg: linear_programming}.
% \end{proof} 

\section{Private synthetic graph generation}\label{sec:PSGG}
Next we provide a precise definition of the (attributed) random graph constructions and their distributions. Furthermore, we give the an algorithm for the joint generation of a network constructed from the data and a private synthetic graph generation; we show that the resulting graphs have the desired distribution.  

\subsection{Graph model}\label{ssec: graph model}
Let $\mcx = \{x_1,\ldots,x_n\} \subset\Omega$ be a data set of attributes and $\kappa:\Omega\times\Omega \to [0,1]$ a symmetric edge connection function; examples include Chung-Lu models and graphon models (see \citet{matias2014modeling} for a survey). Assume that $\kappa$ is Lipschitz continuous with Lipschitz constant $L_{\kappa}$ in both components.
We shall compare the distribution of a graph based on the true attributes $\mcx$ with the distribution of a graph based on the attributes in a subset of $\mcy$ sampled with respect to the private synthetic attribute measure~$\hnu$, see Section~\ref{ssec: Algorithm 1}. 

We start by modeling the ``true'' graph $(\Eta,\Tau)$, that is, the graph constructed from the true input data: We describe the vertices by a Poisson point process~$\Eta$ on $\mcx\times [0,1] \subset \Omega\times [0,1]$ that has (attribute) intensity measure $\mu_n = \frac{a}{n}\sum_{i=1}^n \delta_{x_i}$ for some $a>0$. This corresponds to drawing $N\sim\text{Poi}(a)$ many vertices uniformly from the true data set $\mcx$ and adding a uniform identifier to each vertex, i.e. $\Eta = \sum_{i=1}^N \delta_{(X_i,U_i)}$ with $N \sim \text{Poi}(a)$ as well as $X_i\sim \mathcal{U}(\mcx)$ and $U_i\sim\mcu([0,1])$ for $i\in [N]$. Note that each vertex $(X_i,U_i)$ is unique a.s. as it consists of a (possibly) non-unique attribute $X_i$ and a unique identifier $U_i$. In what follows we often view $(X_i,U_i)\in\Eta$ as vertex $U_i$ marked with attribute $X_i$. In particular, we draw edges between the identifiers $U_i$ with a probability that dependents only on the attributes $X_i$. Furthermore, we measure the distance of two marked points by their attribute distance, i.e. $d_{\Omega\times[0,1]}((x,u),(y,v)) := d_{\Omega}(x,y)$. 

Given $\Eta$, two attributed vertices $(X_i,U_i),(X_i,U_j)\in\Eta$ are connected with probability $\kappa(X_i,X_j)$ independently of other edges.  This yields an edge process $\Tau = \sum_{\substack{i,j=1\\i\neq j}}^N K_{ij}^{\Tau} \delta_{\{U_i,U_j\}}$ for $K_{ij}^{\Tau} \overset{i.i.d.\,}{\sim} \text{Ber}(\kappa(X_i,X_j))$, $i,j\in [N]$. We denote the corresponding distribution of $(\Eta,\Tau)$ by $P^{\mu^a_{\mcx}}\otimes Q^{\kappa}$.

We construct the private synthetic graph $(\Xi,\Sigma)$ on $\mcy\times[0,1]$ with $\mcy = \{y_1,\ldots,y_m\}\subset\Omega$ analogously, as follows. The vertices are given  by a Poisson point process $\Xi = \sum_{i=1}^M \delta_{(Y_i,V_i)}$ where $M\sim\text{Poi}(b)$ for some $b>0$, $V_i\sim \mcu([0,1])$ and $Y_i\in\mcy$ is chosen with respect to $\hat{\nu}$. Furthermore, we define the edge process~$\Sigma$ by $\Sigma = \sum_{\substack{i,j=1\\i\neq j}}^M K_{ij}^{\Sigma} \delta_{\{V_i,V_j\}}$ for $K_{ij}^{\Sigma} \overset{i.i.d.\,}{\sim} \text{Ber}(\kappa(Y_i,Y_j))$. This yields a random attributed graph $(\Xi,\Sigma)\sim P^{\mu^b_{\mcy}}\otimes Q^{\kappa}$. 
% \vspace{-0.2in}
\subsection{The private synthetic graph generator}
We can generate private synthetic graphs by drawing realizations of the private synthetic random graph model described in Section~\ref{ssec: graph model}. The procedure is formally described in Algorithm~\ref{alg: graph generation}. %Note that we can adapt this algorithm to generate ``true'' graphs as realizations of the ``true'' random graph model by ommitting the PSMM step and replacing $\hnu$ by $\mu_{\mcx}:= \frac{1}{a}\mu_n = \frac{1}{n}\sum_{i=1}^n \delta_{x_i}$.
\begin{algorithm}[!ht]			
\caption{Private Synthetic Graph Generation (PSGG)} \label{alg: graph generation}
\begin{algorithmic}
\State {\textbf{Input:}}
 true data $\mcx=(x_1,\ldots,x_n) \in \Omega^n$,  partition $(\Omega_1,\dots,\Omega_m)$ of $\Omega$, privacy parameter $\epsilon >0$, noise distribution $P_{\Lambda}$, expected graph sizes $a$ and $b$, edge probability function $\kappa$.

\begin{description}
  \item[TV-PSMM] Apply Algorithm~\ref{alg: PSMM} to obtain the probability measure $\hnu$ on $\mcy=\{y_1,\ldots,y_m\}$ describing the distribution of private synthetic data.
  \item[Sample graph size] Sample $K_N\sim \text{Poi}(a-(a\wedge b))$, $K_M\sim \text{Poi}(b-(a\wedge b))$ and $L\sim \text{Poi}(a \wedge b)$ and set $N = L + K_N$ and $M = L+K_M$.
  \item[Create common vertex counts] For $j\in [L]$ 
  sample $(Z^j_1,\ldots,Z^j_m)\in\{0,1\}^m$, according to 
\begin{align*}
    \Prob((Z^j_1,\ldots,Z^j_m) = (l_1,\ldots,l_m)) =\begin{cases}
        \frac{n_k}{n}\wedge \hat{\nu}_k\quad &\text{if $l_k = 1$}\\
         1- \sum_{k\in[m]} \frac{n_k}{n}\wedge \hat{\nu}_k \quad &\text{if $\sum_{k\in [m]}l_k = 0$}
    \end{cases}
\end{align*}
for $l_1,\ldots,l_m\in\{0,1\}$ such that $\sum_{k\in[m]} l_k \leq 1$. Set $Z = \sum_{j\in[L]}\sum_{k\in [m]} Z_k^j$. 
\item[Create non-common vertex counts] 
Sample $(M_1,\ldots,M_m) \sim \text{Multin}\big(M-Z, (\hnu_1,\ldots,\hnu_m)\big)$ and
$(N_1,\ldots,N_m) \sim \text{Multin}\bigg(N-Z, \bigg(\frac{n_1}{n},\ldots,\frac{n_m}{n}\bigg)\bigg) $. 
  \item[Create vertices] Set 
  $\Xi = \sum_{k\in[m]}\sum_{j\in [Z_k+M_k]} \delta_{(y_k,V_{kj})},$ with i.i.d.\,identifiers $V_{kj}\sim\mcu[0,1]$ and $\Eta = \sum_{k\in[m]}\sum_{j\in [Z_k+N_k]} \delta_{(X_{kj},U_{kj})},$ with $X_{kj}\sim\mcu(\mcx\cup \Omega_k)$ and i.i.d.\,identifiers $U_{kj}\sim\mcu[0,1]$. 
  \item[Create common edges] For $k,l\in [m]$ and $i\in [Z_k]$, $j\in[Z_l]$ sample ${K}^{\Tau}_{kilj}$ and ${K}^{\Sigma}_{kilj}$ according to
  \begin{align*}
    \Prob(({K}^{\Tau}_{kilj},{K}^{\Sigma}_{kilj})=(e_1,e_2)) =\begin{cases}
        \kappa(X_{ki},X_{lj})\wedge \kappa(y_{k},y_{l})\quad &\text{if $(e_1,e_2) = (1,1)$}\\
        \kappa(X_{ki},X_{lj})-\kappa(X_{ki},X_{lj})\wedge \kappa(y_{k},y_{l})\quad &\text{if $(e_1,e_2) = (1,0)$}\\
        \kappa(y_{k},y_{l})-\kappa(X_{ki},X_{lj})\wedge \kappa(y_{k},y_{l})\quad &\text{if $(e_1,e_2) = (0,1)$}\\
        1-\kappa(X_{ki},X_{lj})\vee \kappa(y_{k},y_{l})\quad &\text{if $(e_1,e_2) = (0,0)$}\\
    \end{cases}.
\end{align*}
  \item[Create common edges] 
 For $i>Z_k$ or $j>Z_l$ sample ${K}^{\Sigma}_{kilj} \sim \text{Ber}(\kappa(y_k,y_l))$ and ${K}^{\Tau}_{kilj} \sim \text{Ber}(\kappa(X_{ki},X_{lj}))$. Set
  $$\Sigma = \sum_{k,l\in[m]}\sum_{i=1}^{M_k}\sum_{j=1}^{M_l} K^{\Sigma}_{kilj} \delta_{\{V_{ki},V_{lj}\}}\quad\text{ and }\quad \Tau = \sum_{k,l\in[m]}\sum_{i=1}^{N_k}\sum_{j=1}^{N_l} K^{\Tau}_{kilj} \delta_{\{U_{ki},U_{lj}\}}.$$
\end{description}

\State {\textbf{Output:}} Private synthetic graph $(\Xi,\Sigma)$ and ``true'' graph $(\Eta,\Tau)$. %synthetic data 
  %$\widehat{\mcy} = (y_1,\ldots,y_{m'}) \in \Omega^{m'}$ for some integer $m'$.
\end{algorithmic}
\end{algorithm}
The following result guarantees that Algorithm~\ref{alg: graph generation} generates suitable graph samples. The full proof is given in Appendix \ref{app:sec:main}.
\begin{theorem}\label{thm: Algorithm 3 distribution}
    The graphs $(\Xi,\Sigma)$ and $(\Eta,\Tau)$ obtained by Algorithm~\ref{alg: graph generation} have distributions $P^{\mu^b_{\mcy}}\otimes Q^{\kappa}$ and $P^{\mu^a_{\mcx}}\otimes Q^{\kappa}$, respectively.
\end{theorem}
\textbf{Sketch of the proof of Theorem~\ref{thm: Algorithm 3 distribution}}
We show that the vertex counts $(M_1,\ldots,M_m)$ and $(N_1,\ldots,N_m)$ constructed in Algorithm~\ref{alg: graph generation} have multinominal distributions and thus the correct marginal distributions. This implies that the vertex processes constructed in Algorithm~\ref{alg: graph generation} define a coupling of the vertex measures of the graph models introduced in Section~\ref{ssec: graph model}. Furthermore, the construction of the edge processes corresponds to a maximal coupling of the corresponding Bernoulli random variables with suitable parameters. Hence, the graphs constructed in Algorithm~\ref{alg: graph generation} have the claimed marginal distributions. \medskip
%\leoni{Proof corresponds to Step 1+2 of original Theorem.}

% Inspired by Proposition 5 in \citet{he2023}, we can show the $\epsilon$-differential privacy under some noise distribution $P_{\Lambda}$.
% \begin{proposition}\label{prop:DP-gaurantee}
%  If $P_\Lambda=\mathrm{Lap}_{\mathbb{Z}}(1/\epsilon)$ is the discrete Laplace distribution, then the Algorithm~\ref{alg: graph generation} is $\epsilon$-differential private in vertex-level. 
% \end{proposition}
% \subsection{Total variation complexity}
% We aim to control the distributional differences of random graphs that are generated based on random draws from the true and synthetic data, respectively. This distributional comparison is based on suitable couplings of the random graphs. In particular, to bound the distributional differences with respect to the fused Gromov--Wasserstein distance, we need to control the total variation distance of the relative true counts~$n_i/n$ and the synthetic count probability~$\hnu_i$.

\begin{figure}
    \centering
    \includegraphics[width=0.24\linewidth]{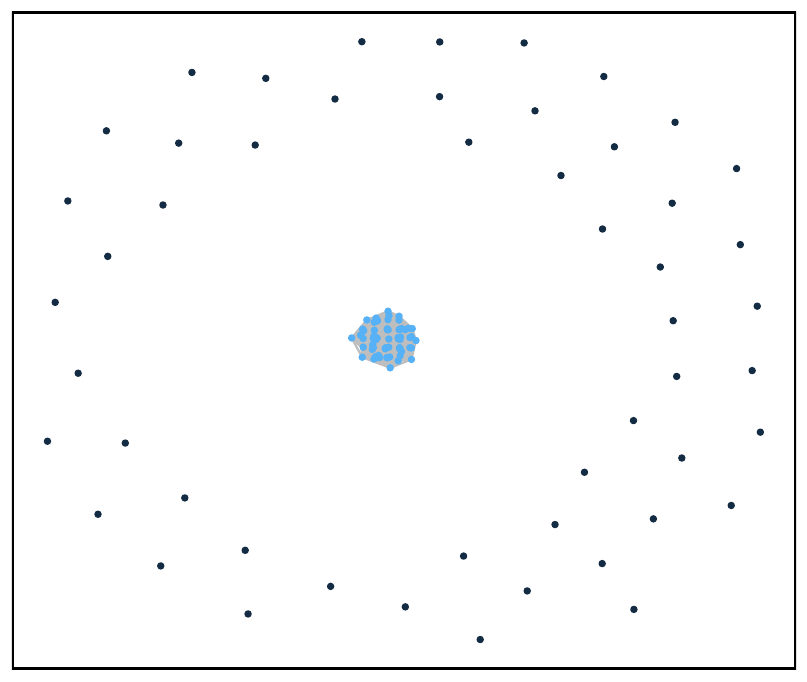}
    \includegraphics[width = 0.24\linewidth]{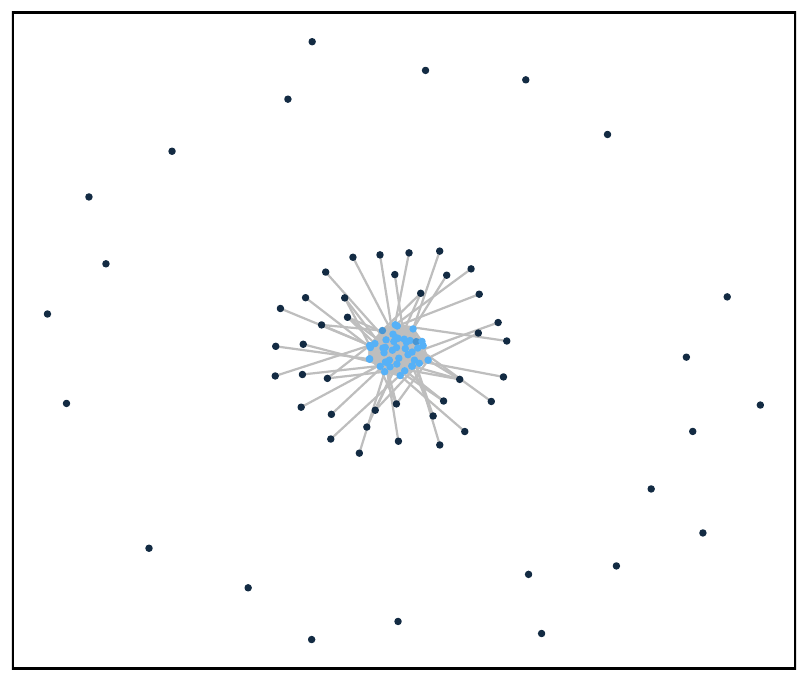}
    \includegraphics[width = 0.24\linewidth]{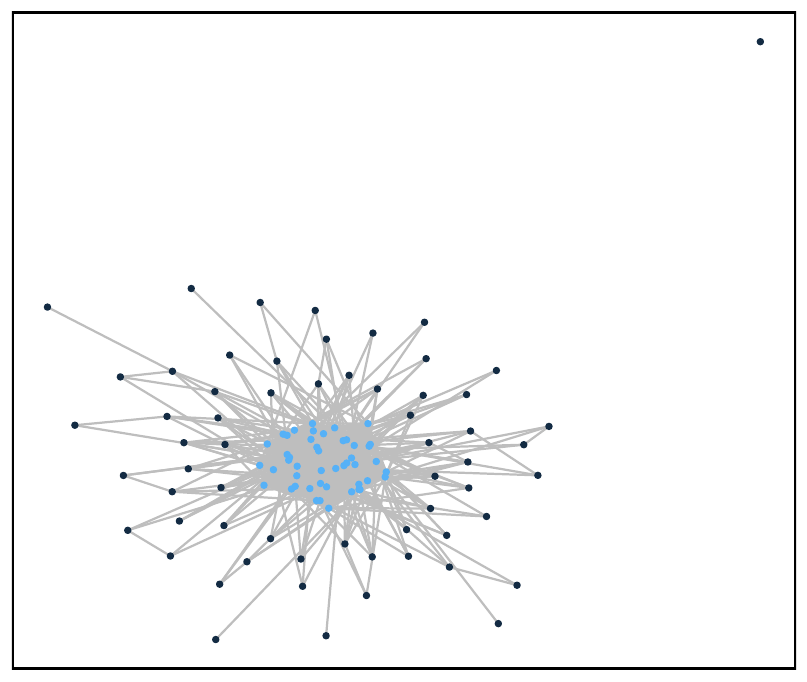}
    \includegraphics[width = 0.24\linewidth]{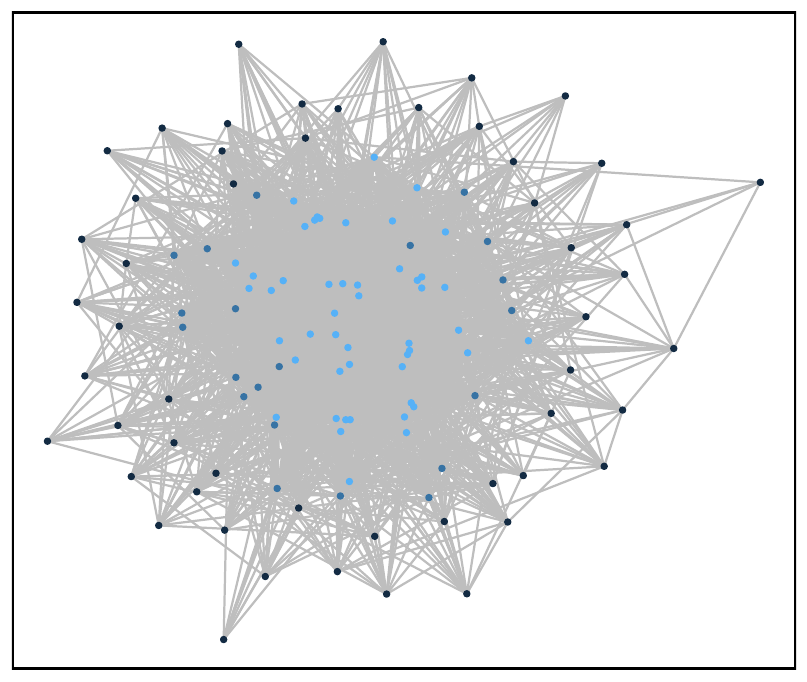}
    \caption{Graphs of Chung-Lu type generated with Algorithm~\ref{alg: graph generation} on $\Omega = [0,1]$ with $n=1000$, $m = \lceil\sqrt{\epsilon n}\rceil$ and $a=b=100$. Graph~1 shows a realization of the ``true'' graph with $\mcx = \{0,\ldots,0,1,\ldots,1\}$. Graph~2-4 are realizations of the private synthetic graph generated with Laplacian noise $\text{Lap}_{\Z}(1/\epsilon)$ for $\epsilon\in\{1,0.1,0.01\}$. The vertex color visualizes the attribute, where small attributes correspond to dark colors.}
    \label{fig: chung lu type graph}
\end{figure}

We present an example of this graph generation is Figure~\ref{fig: chung lu type graph} for increasing privacy parameter $\epsilon$. The graphs are generated using attributes on $\Omega = [0,1]$ and $\kappa(x,y) = xy$, leading a random graph model of Chung-Lu type, where vertices with high weights/attributes are more likely to form edges than vertices with small weights. %\gr{CL would take the product; this is more of graphon type}

% \subsection{Distribution of Noise} \label{ssec: noise}
% \leoni{We can deal with general i.i.d.\, noise $\Lambda$. For the upper bound we need to calculate $\E\abs{\Lambda}$.}
% \leoni{Maybe we can connect this section with some privacy content.}
% \todoa{I consider $\mathcal{N}_{\mathbb{Z}}(0,\frac{n^2}{\epsilon^2})$ to be different from \cite{he2023}}
% \todoa{What is the application of Section 3.3? }
% case 1: $\Lambda \sim \text{Lap}_{\mathbb Z}(1/\epsilon)$\\
% Then by \cite[Chapter~2]{inusah2006}, we have for $p = e^{-\epsilon}$ that 
% \begin{align*}
%     \E\abs{\Lambda} = 2\frac{1+p}{1-p}\sum_{k=1}^{\infty} k p^k = 2\frac{1+p}{1-p}\frac{p}{(1-p)^2} = 2\frac{p}{1-p^2} = 2\frac{e^{-\epsilon}}{1-e^{-2\epsilon}}.
% \end{align*}
% where we used that
% \begin{align*}
%     &\sum_{k=1}^{\infty} k p^k = p + p \sum_{k=1}^{\infty} (k+1) p^k 
%     =p + p \sum_{k=1}^{\infty} k p^k + p\sum_{k=1}^{\infty} p^k 
%     =p + p \sum_{k=1}^{\infty} k p^k + \frac{p}{1-p}-p; \end{align*}  
%     re-arranging this equation gives
%     \begin{align*}
%      \sum_{k=1}^{\infty} k p^k =  \frac{p}{(1-p)^2}.
% \end{align*}

\section{Theoretical guarantees on utility}\label{sec:Main}
In this section we give theoretical results  for the PSGG algorithm which show conditions under which the differentially private synthetic network distribution is close to the  distribution of the ``true'' graph. 
%We analyze the effect of the TV-PSMM algorithm on the generation of graphs in 
Algorithm~\ref{alg: graph generation}. In particular, %\gr{try to avoid more precisely}\todoa{Done.} 
we study the differences between the ``true'' graph~$(\Eta,\Tau)$ that was constructed using the true data $\mcx$, and the private synthetic graph $(\Xi,\Sigma)$ that was constructed using the private synthetic measure $\hnu$. In the first part we consider the accuracy of Algorithm~\ref{alg: graph generation} while the second part analyzes the difference in distribution. %Our results are obtained using two different approaches, which are inspired by coupling method and Stein's method \cite{schuhmacher2024}. 
% As a first approach, we employ the coupling method to compare the graphs in a direct way, while for the second approach we apply results from  on the comparison of random graph distributions. 
All proofs details are deferred to App.~\ref{app:sec:main}. 
\subsection{Accuracy of Algorithm~3}\label{ssec: coupling approach}
 %\gr{perhaps rather: Bounds on the expected FGW distance?}
In the following result we control the accuracy measure, i.e. we bound the expected distance between the true attributed random graph~$(\Eta,\Tau)$ and the private synthetic attributed random graph~$(\Xi,\Sigma)$ constructed by Algorithm~\ref{alg: graph generation} with respect to the FGW distance, see Section~\ref{ssec: FGW}. That is, we study $\E(d_{FGW}(\eta^{(\Eta,\Tau)},\eta^{(\Xi,\Sigma)}))$, where $\eta^{(\Eta,\Tau)}$ and $\eta^{(\Xi,\Sigma)}$ are random measures that capture the (attributed) graph information of $(\Eta,\Tau)$ and $(\Xi,\Sigma)$, respectively. 
%The proof is based on the construction of a coupling of the two random graphs, i.e. on a suitable definition of the two random graphs on a common probability space. 
We recall the notion of diameter; $diam(\Omega) = \sup_{x,y \in \Omega} d(x,y).$

\begin{theorem}\label{thm: result coupling}
Let $(\Eta,\Tau)$ and $(\Xi,\Sigma)$ be the ``true'' and private synthetic random graphs of expected size $a>0$ and $b>0$, respectively, constructed by Algorithm~\ref{alg: graph generation}. % using a probability measure $\hnu$ created by Algorithm~\ref{alg: PSMM} with random noise $\Lambda$. 
Then 
\begin{align*}
    \E\big(d_{FGW}(\eta^{(\Eta,\Tau)},\eta^{(\Xi,\Sigma)})\big) %\leq \tilde{C}_1 \big(\max_{k\in[m]}\mathrm{diam}(\Omega_k)\big)  \sum_{k\in[m]}  \frac{n_k}{n}\wedge \hat{\nu}_k+2 \tilde{C}_2\frac{m}{n} \E\abs{\Lambda}
    &\leq \tilde{C}_1\,\max_{k\in[m]}\mathrm{diam}(\Omega_k) \frac{1}{1+\frac{a\wedge b}{\abs{a-b}}\1_{\{\abs{a-b}>0\}}} + 4\tilde{C}_2 \frac{m}{n} \E(\abs{\Lambda})\\
    &+ \tilde{C}_2 \bigg(1 - \frac{1}{\big(1+\frac{a\wedge b}{\abs{a-b}}\big)^2}\bigg) \1_{\{\abs{a-b}>0\}},
\end{align*}
where $\tilde{C}_1 = 1-\alpha+\alpha (2\,C\,L_{\kappa}) $ and $\tilde{C}_2 =  (1-\alpha)\mathrm{diam}(\Omega)+\alpha \min\{C,2\,C\,L_{\kappa} \mathrm{diam}(\Omega)\} $ with $\alpha\in [0,1]$ and $C>0$ parameters of the FGW metric and $L_{\kappa}$ the Lipschitz constant of edge connection function~$\kappa$.%\gr{we should introduce the trade-off parameter earlier} the trade-off parameter and the bound of the edge cost\gr{this part seems to be interpretation of the bound and should then go outside the theorem} for the FGW metric, respectively. 
\end{theorem}
%The key issue of the proof\todoa{which proof?} is that the different sampling for the attributes (according to $\mcx$ and $\hnu$, resp.) will lead to different proportions of vertices in each cell $\Omega_i$. We approach this by constructing a coupling of the vertex counts. 
In what follows we sketch the idea of the proof and refer to App.~\ref{app:sec:main} for details. Note that the construction of the graphs in Algorithm~\ref{alg: graph generation} allows us to significantly reduce the costs of the estimates.
\medskip\\
{\bf Sketch of the proof of Theorem~\ref{thm: result coupling}:} 
%\leoni{The statement is now reformulated. The proof of this statement is only Step 3 of the old proof. }
For two pairs of vertices $(X_i,Y_j)$ we can bound the attribute cost by $\mathrm{diam}(\Omega)$ and the edge cost by $\min\{C,2\,C\,L_{\kappa} \mathrm{diam}(\Omega)\}$ using the Lipschitz property of $\kappa$ for the second term. Under the assumption that $(X_i,Y_j)$ are in the same cell, this bound can be improved to $\max_{k\in[m]}\mathrm{diam}(\Omega_k)$ for the vertex part and $2\,C\,L_{\kappa} \max_{k\in[m]}\mathrm{diam}(\Omega_k)$ for the edge part; details are found in Appendix \ref{app:sec:main}.
We then obtain the result by conditioning on the (minimal) number $Z$ of vertices $(X_i,Y_i)$ that are in a common cell.\medskip

We consider a special case of Theorem~\ref{thm: result coupling} to give a more thorough discussion of the obtained bound. We restrict to discrete Laplacian noise $\Lambda\sim\text{Lap}_{\Z}(1/\epsilon)$ which yields $\E\abs{\Lambda}  = 2\frac{e^{-\epsilon}}{1-e^{-2\epsilon}}$, see \cite{inusah2006}, and set $\Omega = [0,1]^d$ to be the $d$-dimensional unit cube. Moreover, we let $a=b$ and choose the sample size $m= \lceil f_n(\epsilon)n\rceil$ for a function $f_n:\R_+\to \R_+$.  Under these assumptions the bound obtained above simplifies to the following result.
\begin{corollary} \label{cor: special case coupling bounds}%[Special case]
Let $m = \lceil f_n(\epsilon)n\rceil$ and consider graphs generated by Algorithm~\ref{alg: graph generation} using discrete Laplacian noise and a partition $(\Omega_k)_{k=1}^m$ which satisfies $\mathrm{diam}(\Omega_k)\asymp m^{-1/d}$. Assume that the graphs have %(in expectation) 
the same expected size $a=b$ with attributes in $\Omega = [0,1]^d$. %Furthermore, assume that the probability measure $\hnu$ is created by Algorithm~\ref{alg: PSMM} using discrete Laplacian noise and a partition $(\Omega_k)_{k=1}^m$ satisfying $\mathrm{diam}(\Omega_k)\asymp m^{-1/d}$. 
Then Theorem~\ref{thm: result coupling} simplifies to
\begin{align} \label{eq:fn}
    \E\big(d_{FGW}(\eta^{(\Eta,\Tau)},\eta^{(\Xi,\Sigma)})\big) 
    &\leq \tilde{C}_1 m^{-1/d} +4 \tilde{C}_2 f_n(\epsilon) \frac{e^{-\epsilon}}{1-e^{-2\epsilon}}.
\end{align}
\end{corollary}
A reasonable choice of the function~$f_n$ is essential to obtain good convergence rates and we refer to Section~\ref{ssec: choice of parameters} for an analysis.

% \begin{remark}\label{re: trade off coupling bound}
% We obtain a trade-off for the choice of $f_n$. More precisely, to obtain fast convergence (to zero) in the first term, we would like $f_n(\epsilon)$ to go to zero slowly. On the other hand, the second term decreases faster the faster $f_n(\epsilon)$ goes to zero. We derive an optimal decay for $f_n$ as follows. 
% First, note that ${\epsilon\,e^{-\epsilon}}/({1-e^{-2\epsilon}})\xrightarrow{\epsilon\to 0}1/2.$
% In particular, the second %\gr{third term? What about the constant term?}\leoni{Typo. Only two terms.}
% term in \eqref{eq:fn} asymptotically behaves as $f_n(\epsilon)/\epsilon$. On the other hand, we chose $m = \lceil f_n(\epsilon)n\rceil$ which implies that the first term asymptotically behaves like $\big(f_n(\epsilon)n\big)^{-1/d}.$ Due to the trade-off between these two terms, an optimal decay for $f_n$ can be obtained by assuming the same speed of convergence for both terms. This corresponds to setting 
% \begin{align*}
%      \frac{f_n(\epsilon)}{\epsilon} \asymp \big(f_n(\epsilon)n\big)^{-1/d} \iff f_n(\epsilon ) \asymp \epsilon^{\frac{d}{d+1}}\,n^{-\frac{1}{d+1}}.
% \end{align*}
% In particular, under the assumptions of Corollary~\ref{cor: special case coupling bounds} we obtain
% \begin{align*}
%     \E\big(d_{FGW}(\eta^{(\Eta,\Tau)},\eta^{(\Xi,\Sigma)})\big) 
%     &\leq \tilde{C}_1 \big(\epsilon\,n\big)^{-1/(d+1)} +2 \tilde{C}_2 \big(\epsilon\,n\big)^{-1/(d+1)} = (\tilde{C}_1 + 2\tilde{C}_2) \big(\epsilon\,n\big)^{-1/(d+1)}.
% \end{align*}
% \end{remark}
\begin{remark}
A special case of Theorem~\ref{thm: result coupling} can be obtained by setting the edge probability function $\kappa \equiv 0$. Then we a.s. have no edges and the random graphs constructed as in Algorithm~\ref{alg: graph generation} can be described by their vertex processes $\Eta$ and $\Xi$. The FGW distance then simplifies to the 1-Wasserstein distance. In particular, we can bound the accuracy of Algorithm~\ref{alg: graph generation} for vertex generation with the results obtained in Theorem~\ref{thm: result coupling} by setting $\alpha = 0$.
\end{remark}

\subsection{Bounds on the difference between the graph distributions} \label{ssec: stein approach}
In the following we analyze and measure the differences between the distribution $(P^{\mu^a_{\mcx}}\otimes Q^{\kappa})$ of the ``true'' random graph $(\Eta,\Tau)$ and the distribution $(P^{\mu^b_{\mcy}}\otimes Q^{\kappa})$ of the private synthetic random graph $(\Xi,\Sigma)$. To measure the distance of random graph distributions, we consider the Wasserstein distance with respect to the FGW distance as an underlying metric. More precisely, we define 
\begin{equation} \label{eq: integral metric}
    d_{\mcf}(P^{\mu^b_{\mcy}}\otimes Q^{\kappa} ,P^{\mu^a_{\mcx}}\otimes Q^{\kappa}) = \sup\limits_{f\in\mcf} \big\vert \E(f(\Xi,\Sigma)) - \E(f(\Eta,\Tau))\big\vert,
\end{equation}
where $\mcf$ is the set of test functions $f:\G \to \R$ that are Lipschitz with respect to the FGW distance, see Section~\ref{ssec: FGW}. As a direct consequence of this Lipschitz property and Theorem~\ref{thm: Algorithm 3 distribution}, the results obtained in Section~\ref{ssec: coupling approach} can be in particular used to bound the above distance.
%Due to the Lipschitz property of functions $f\in\mcf$, Theorem~\ref{thm: Algorithm 3 distribution} and Theorem~\ref{thm: result coupling} imply a bound on the distance of the corresponding graph distributions w.r.t. $d_{\mcf}$, see Section~\ref{ssec: FGW}.
% \begin{corollary)\label{cor: coupling becomes distirbution}
%     Under the assumptions of Theorem~\ref{thm: result coupling} we obtain
%     \begin{align*}
%     d_{\mcf}(P^{\mu^b_{\mcy}}\otimes Q^{\kappa} ,P^{\mu^a_{\mcx}}\otimes Q^{\kappa})
%     &\leq \tilde{C}_1\,\max_{k\in[m]}\mathrm{diam}(\Omega_k) \frac{1}{1+\frac{a\wedge b}{\abs{a-b}}\1_{\{\abs{a-b}>0\}}} + 4\tilde{C}_2 \frac{m}{n} \E(\abs{\Lambda})\\
%     &+ \tilde{C}_2 \bigg(1 - \frac{1}{\big(1+\frac{a\wedge b}{\abs{a-b}}\big)^2}\bigg) \1_{\{\abs{a-b}>0\}}.
% \end{align*}
% %where $\tilde{C}_1 = 1-\alpha+\alpha (2\,C\,L_{\kappa}) $ and $\tilde{C}_2 =  (1-\alpha)\mathrm{diam}(\Omega)+\alpha \min\{C,2\,C\,L_{\kappa} \mathrm{diam}(\Omega)\} $ with $\alpha\in [0,1]$ and $C>0$ parameters of the FGW metric and $L_{\kappa}$ the Lipschitz constant of edge connection function~$\kappa$.
% \end{corollary}

We derive upper bounds for these distances of random graph distributions by adapting general bounds from  \cite{schuhmacher2024}.
%using Stein's method. 
We remark that, in general, it is possible to consider different classes of test functions and thus obtain different types of so called integral probability metrics. Our approach will then work analogously, however, the obtained upper bounds and rates will strongly depend on the choice of test functions.  
\begin{theorem}\label{thm: result Stein}
Let $(P^{\mu^a_{\mcx}}\otimes Q^{\kappa})$ and $(P^{\mu^b_{\mcy}}\otimes Q^{\kappa})$ be distributions of the ``true'' and the private synthetic random graph of expected size $a>0$ and $b>0$, respectively, constructed as in Section~\ref{ssec: graph model} by using a probability measure $\hnu$ created by Algorithm~\ref{alg: PSMM} with random noise $\Lambda$.
Then 
    \begin{align*}
        &d_{\mcf}(P^{\mu^b_{\mcy}}\otimes Q^{\kappa} ,P^{\mu^a_{\mcx}}\otimes Q^{\kappa})\\
        &\leq  \bigg(1+\frac{1}{1+\frac{b}{c}\1_{\{c>0\}}}\bigg) (1-\alpha) \, \max_{k\in[m]}\mathrm{diam}(\Omega_k)+\tilde{C}_2\bigg(1 - \frac{1}{\big(1+\frac{b}{c}\big)^2}\bigg) \1_{\{c>0\}}\\
        &+ 2c_V(a\wedge b)\frac{a\wedge b}{n}\Leb^d(\Omega) \E\abs{\Lambda} + c_E(a\wedge b)\,2L_{\kappa}\max_{k\in[m]}\mathrm{diam}(\Omega_k)^3 (a\wedge b)^2,\\
    \end{align*}
    where
\begin{align*}
   c_V(c) &:= \min \biggl\{ C_{\alpha}, % B^{*}
    \frac{1}{c} \,\bigl(1 +  ({1-e^{-{c}}}) \log^+{c} \bigr)\,C_{\alpha} \biggr\},\\
    c_E(c) &:= \min \biggl\{ 1, \frac{2-e^{-c}}{c} - \frac{1}{c^2}\Bigl(\frac{3}{2} - e^{-c}\Bigr) \biggr\} \; \alpha C
\end{align*}
for $C_{\alpha} := (1-\alpha)\mathrm{diam}(\Omega) + \alpha C$ and $\tilde{C}_2 = (1-\alpha)\mathrm{diam}(\Omega) + \alpha \min\{C,2CL_{\kappa}\mathrm{diam}(\Omega)\}$ %an upper bound of $d_{FGW}$ 
with $\alpha\in [0,1]$ and $C>0$ parameters of the FGW distance and $L_{\kappa}$ the Lipschitz constant of the edge connection function~$\kappa$.% \gr{rephrase?}
\end{theorem}
In what follows, we give a short sketch of the proof for Theorem~\ref{thm: result Stein}; we  refer to App~\ref{app:sec:main} for details. 

\medskip 
%\begin{proof}
% [
\textbf{Sketch of the proof of Theorem~\ref{thm: result Stein}. }%]
 The proof is based on results of \cite[Theorem~4.7]{schuhmacher2024}, which provides upper bounds on the distance of two random (attributed) graph distributions,  derived by Stein's method. The bounds in \cite[Theorem~4.7]{schuhmacher2024} are of total variation type and thus will not vanish in a direct comparison (as the attribute measures have different support). Furthermore, the bound is sensitive to the number of vertices. More precisely, the ``true'' graph has mass $N$ while the private synthetic graph has mass $M$. We approach this issue  by creating intermediate graphs that are supported on whole $\Omega$ and have the same size $N\wedge M$. These intermediate graphs are constructed by resampling the attributes of the first $N\wedge M$ points in each cell while keeping the edge probabilities of the original graph. The difference between the original graphs and their corresponding intermediate graphs can be analyzed using a coupling. %, which is a special instance of the coupling constructed to prove Theorem~\ref{thm: result coupling}. Note that the vertex counts do not change as we resample the attributes in each cell. Furthermore, by construction the edge probabilities are not modified. Thus, we obtain significantly lower bounds compared to the results of Theorem~\ref{thm: result coupling}.
It remains to compare the intermediate graphs. This now can be done by an %direct
application of \citet[Theorem~4.7]{schuhmacher2024}, stated in terms of their so-called {\it GOSPA} metric. %noting that %. The Stein factors are derived for the GOSPA metric. However, 
Under a suitable choice of parameters in the GOSPA metric, the FGW distance is a lower bound for the GOSPA metric, which suffices for our purposes. %In particular, the Stein factors hold in our application. 
Finally, we assemble all results to obtain to obtain the claimed upper bound.  %\hfill $\Box$

%\end{proof}

\begin{corollary} \label{cor: special case Stein}%[Special case]
Consider graphs being distributed according to Section~\ref{ssec: graph model} w.r.t. a probability measure $\hnu$ created by Algorithm~\ref{alg: PSMM} using discrete Laplacian noise and a partition $(\Omega_k)_{k=1}^m$ satisfying $\mathrm{diam}(\Omega_k)\asymp m^{-1/d}$. Furthermore, set $a=b$ and $\Omega = [0,1]^d$. Then Theorem~\ref{thm: result Stein} simplifies to %for the distance between the distributions of the true and synthetic random graph we obtain \gr{explain $\alpha, \epsilon,  L_{\kappa}, C$}
\begin{align} \label{eq: cor special case Stein}
    &d_{\mcf}(P^{\mu^b_{\mcy}}\otimes Q^{\kappa} ,P^{\mu^a_{\mcx}}\otimes Q^{\kappa})
    %&\leq 2 (1-\alpha) \, m^{-1/d}
   % + \frac{2}{n} \,\bigl(1 +  ({1-e^{-{a}}}) \log^+{a} \bigr)\,C_{\alpha} \E\abs{\Lambda} + \biggl(2-e^{-a} - \frac{1}{a}\Bigl(\frac{3}{2} - e^{-a}\Bigr)\biggr)  \; \alpha C\,2L_{\kappa} m^{-3/d} a \notag\\
    \leq 2 (1-\alpha) \, m^{-1/d}
    + \frac{2\bigl(1 +   \log^+{a} \bigr)}{\epsilon n}C_{\alpha} \epsilon\frac{e^{-\epsilon}}{1-e^{-2\epsilon}} + 4 \alpha CL_{\kappa} m^{-3/d} a .
    % &\leq 2 (1-\alpha) \, \max_{k\in[m]}\mathrm{diam}(\Omega_k)+ \frac{1 + \log^+b}{n} C_{\alpha}\E\abs{\Lambda} \\
    % &  + \frac{1}{b}\, \alpha C\,\sum_{i,j=1}^m\Leb^d(\Omega_i)\Leb^d(\Omega_j)\,2L_{\kappa}\max_{k\in[m]}\mathrm{diam}(\Omega_k) \,\frac{n_i}{n}\,\hat{\nu}_j\,b^2\\
    % &\leq 2 (1-\alpha) \, \max_{k\in[m]}\mathrm{diam}(\Omega_k)+ \frac{1 + \log^+b}{n} C_{\alpha}\E\abs{\Lambda} \\
    % &  + \frac{1}{b}\, \alpha C\,\sum_{i,j=1}^m 2L_{\kappa}\big(\max_{k\in[m]}\mathrm{diam}(\Omega_k) \big)^3 \,\frac{n_i}{n}\,\hat{\nu}_j\,b^2\\
    % &\asymp 2 (1-\alpha) \,m^{-1/d} + 2C_{\alpha} \frac{1 + \log^+b}{n} \E\abs{\Lambda}  +  2\alpha C\, L_{\kappa}\, b\,m^{-3/d} 
    %&\overset{a= m^{2/d}}{\leq } 2 (1-\alpha) \, m^{-1/d}
    %+ \frac{2}{\epsilon n} \,\bigl(1 + 2/d\log^+{m} \bigr)\,C_{\alpha} \epsilon\frac{e^{-\epsilon}}{1-e^{-2\epsilon}} + 4 \alpha C\,L_{\kappa} m^{-1/d}  . 
\end{align}
\end{corollary}
Similar to the results in Section~\ref{ssec: coupling approach}, a reasonable choice of the (expected) size $a = a_n$ of the random graph  and the partition size $m= \lceil f_n(\epsilon)n\rceil$ are essential to obtain good convergence rates, as we discuss in the next section.

\subsection{Choice of parameters}\label{ssec: choice of parameters}

In practice, we often are given a dataset $\mcx$ of size $n$ and a privacy level $\epsilon$ and have to make suitable choices for the expected size of the graphs $a$ and the partition size $m= \lceil f_n(\epsilon)n\rceil$. Here  we discuss a suitable choice of these parameters using the rates obtained by Theorem~\ref{thm: result coupling} and Theorem~\ref{thm: result Stein}. Again, we restrict to graphs having the same expected size $a=b$ with attributes on the unit cube $\Omega = [0,1]^d$ and assume that the graphs are generated by Algorithm~\ref{alg: graph generation} using discrete Laplacian noise $\Lambda\sim \text{Lap}_{\Z}(1/\epsilon)$. Furthermore, we let $m = \lceil f_n(\epsilon) n \rceil$ for some function $f_n$ and remark that for any $\epsilon\geq 0$
$$\epsilon\,\frac{e^{-\epsilon}}{{1-e^{-2\epsilon}}}\leq 1/2.$$ 
Then Corollary~\ref{cor: special case coupling bounds} yields 
\begin{equation}\label{eq: coupling bound special case}
    d_{\mcf}(P^{\mu^b_{\mcy}}\otimes Q^{\kappa} ,P^{\mu^a_{\mcx}}\otimes Q^{\kappa})
    \leq \E\big(d_{FGW}(\eta^{(\Eta,\Tau)},\eta^{(\Xi,\Sigma)})\big) 
    \leq \tilde{C}_1 (f_n(\epsilon)n)^{-1/d} +2 \tilde{C}_2 \frac{f_n(\epsilon)}{\epsilon},
\end{equation}
where the first inequality follows from the definition of $d_{\mcf}$, see Section~\ref{ssec: stein approach}; while Corollary~\ref{cor: special case Stein} states 
\begin{align} \label{eq: stein bound special case}
    &d_{\mcf}(P^{\mu^b_{\mcy}}\otimes Q^{\kappa} ,P^{\mu^a_{\mcx}}\otimes Q^{\kappa}) \\ \notag
    &\leq 2 (1-\alpha) \, m^{-1/d}
    + \frac{2\bigl(1 +   \log^+{a} \bigr)}{\epsilon n}C_{\alpha} \epsilon\frac{e^{-\epsilon}}{1-e^{-2\epsilon}} + 4 \alpha CL_{\kappa} m^{-3/d} a .
\end{align}
The speed of convergence depends on the choice of $f_n$ as well as the expected size of the graphs~$a$. We start with a suitable choice of graph size. Consider the bound in \eqref{eq: stein bound special case}.
On the one hand, we would like to choose the expected graph size $a = a_n$ as large as possible to have a good sample size for our graphs. On the other hand, \eqref{eq: cor special case Stein} yields better bounds the smaller $a$ is. 
As the first term in \eqref{eq: cor special case Stein} does not depend on $a$, an optimal choice of $a$ is such that the first and third terms in \eqref{eq: cor special case Stein} have the same speed of convergence. Note that we do not consider the second term here as due to the log-term the influence of $a$ is negligible compared to the third term. Then 
$$m^{-1/d} \asymp a m^{-3/d} \iff a = m^{2/d}.$$
In particular, using that $m= \lceil f_n(\epsilon)n\rceil$ Inequality \eqref{eq: stein bound special case} turns to 
\begin{align} \label{eq: stein bound special case 2}
    &d_{\mcf}(P^{\mu^b_{\mcy}}\otimes Q^{\kappa} ,P^{\mu^a_{\mcx}}\otimes Q^{\kappa}) \leq \big(\! 2 (1-\alpha) +4 \alpha CL_{\kappa}\!\big) (f_n(\epsilon)n)^{-\frac{1}{d}}
    + C_{\alpha} (1\!+2\log^+\!(f_n(\epsilon)n)\!)(\epsilon n)^{-1} .
\end{align}
It remains to find a suitable choice of $m= \lceil f_n(\epsilon)n\rceil$ by deriving an optimal decay of $f_n(\epsilon)$ as $\epsilon\to 0$. Note that in both, \eqref{eq: coupling bound special case} and \eqref{eq: stein bound special case 2}, we obtain a trade-off for the choice of $f_n$. More precisely, to obtain fast convergence (to zero) in the first terms, we would like $f_n(\epsilon)$ to go to zero slowly. On the other hand, the second terms decrease faster the faster $f_n(\epsilon)$ goes to zero. 
However, note that due to the logarithmic expression the observed trade-off in \eqref{eq: stein bound special case 2} is weaker than the trade-off in \eqref{eq: coupling bound special case}. In particular, we focus on \eqref{eq: coupling bound special case} and derive an optimal decay for $f_n$ as follows.

Due to the trade-off between the two terms in \eqref{eq: coupling bound special case}, an optimal decay for $f_n$ can be obtained by assuming the same speed of convergence for both terms in \eqref{eq: coupling bound special case}. This corresponds to setting 
\begin{align*}
     \frac{f_n(\epsilon)}{\epsilon} \asymp \big(f_n(\epsilon)n\big)^{-1/d} \iff f_n(\epsilon ) \asymp \epsilon^{\frac{d}{d+1}}\,n^{-\frac{1}{d+1}}.
\end{align*}
In particular, under the above assumptions \eqref{eq: coupling bound special case} and thus Corollary~\ref{cor: special case coupling bounds} simplify to 
\begin{align*}
    d_{\mcf}(P^{\mu^b_{\mcy}}\otimes Q^{\kappa} ,P^{\mu^a_{\mcx}}\otimes Q^{\kappa}) \leq \E\big(d_{FGW}(\eta^{(\Eta,\Tau)},\eta^{(\Xi,\Sigma)})\big) 
    &\leq %\tilde{C}_1 \big(\epsilon\,n\big)^{-1/(d+1)} +2 \tilde{C}_2 \big(\epsilon\,n\big)^{-1/(d+1)} = 
    (\tilde{C}_1 + 2\tilde{C}_2) \big(\epsilon\,n\big)^{-\frac{1}{d+1}}.
\end{align*}
Similarly, \eqref{eq: stein bound special case 2} and thus Corollary~\ref{cor: special case Stein} turn to 
\begin{align*}
    &d_{\mcf}(P^{\mu^b_{\mcy}}\otimes Q^{\kappa} ,P^{\mu^a_{\mcx}}\otimes Q^{\kappa}) \leq \big( 2 (1-\alpha) +4 \alpha CL_{\kappa}\big) (\epsilon n)^{-\frac{1}{d+1}}
    + C_{\alpha} \bigg(1+\frac{2}{d+1}\log^+(\epsilon n)\bigg)(\epsilon n)^{-1}.
\end{align*}
In particular, with the above choice of $f_n$ and $a$ we obtain the same speed of convergence for both approaches. Note however, that we optimized $f_n$ w.r.t. the results obtained in Section~\ref{ssec: coupling approach}. Hence, for a different choice of $f_n$ the results stated in Section~\ref{ssec: stein approach} can yield improved rates.

We end this Section by giving a table with precise upper bounds, Table \ref{tab: explicit bounds} and refer to Appendix~\ref{app:sec:main} for explicit upper bounds in different parameter settings. There is no uniformly better bound Corollary~\ref{cor: special case coupling bounds} gives smaller bounds than Corollary~\ref{cor: special case Stein} when $\epsilon$ is large. Both bounds decrease when $n$ increases. 

\begin{table}[ht]
    \centering
    \begin{filecontents*}{Tables/tableBoundCompOptfnAlpha12.csv}
"";"as.character.eps.1..";"X100";"X100.1";"X1000";"X1000.1";"X10000";"X10000.1"
"X.";"eps";"n = 100";"n = 100";"n = 1000";"n = 1000";"n = 10000";"n = 10000"
"X1";"1";"0.751";"0.681";"0.349";"0.304";"0.162";"0.14"
"X2";"0.1";"1.599";"1.602";"0.751";"0.681";"0.349";"0.304"
"X3";"0.01";"3.061";"3.721";"1.599";"1.602";"0.751";"0.681"
\end{filecontents*}
\pgfplotstabletypeset[
        col sep=semicolon,
        string type,
        column type=c,
        string replace*={"}{},
        header=false,
        every head row/.style={ 
            output empty row,
            before row={%
                \hline
                \multicolumn{1}{|c}{}&
                \multicolumn{1}{c|}{}&
                \multicolumn{6}{c|}{$n$}\\[0.5ex]
                \cline{3-8}
                \multicolumn{1}{|c}{}&
                \multicolumn{1}{c|}{}&
                \multicolumn{2}{c|}{100}&
                \multicolumn{2}{c|}{1000}&
                \multicolumn{2}{c|}{10.000}\\[0.5ex]
                \cline{3-8}
                \multicolumn{1}{|c}{}&
                \multicolumn{1}{c|}{}&
                \multicolumn{1}{c|}{Cor~\ref{cor: special case coupling bounds}}&
                \multicolumn{1}{c|}{Cor~\ref{cor: special case Stein}}&
                \multicolumn{1}{c|}{Cor~\ref{cor: special case coupling bounds}}&
                \multicolumn{1}{c|}{Cor~\ref{cor: special case Stein}}&
                \multicolumn{1}{c|}{Cor~\ref{cor: special case coupling bounds}}&
                \multicolumn{1}{c|}{Cor~\ref{cor: special case Stein}}\\
                \hline
            }
        },
        create on use/newCol/.style={
            create col/set list={" "," "," ",$\epsilon$, " "}
        },
        skip rows between index={0}{2},
        columns/newCol/.style={
            string type,
            column type/.add={|}{}
        },
        columns={newCol, [index]1, [index]2, [index]3, [index]4, [index]5, [index]6, [index]7},
        every last row/.style={after row=\hline},
        every odd column/.style={column type/.add={}{|}},
        every even column/.style={column type/.add={}{|}},
        every row/.style={before row={\rule{0pt}{2.5ex}}}
    ]{Tables/tableBoundCompOptfnAlpha12.csv}
    \caption{Explicit bounds obtained in Corollary~\ref{cor: special case coupling bounds} and Corollary~\ref{cor: special case Stein} assuming discrete Laplacian noise and $\Omega = [0,1]^2$, $\alpha = 1/2$ as well as $C = L_{\kappa} = 1$. Furthermore, $m = \lceil f_n(\epsilon)n\rceil$ with $f_n$ optimal and $a$ optimal w.r.t. Section~\ref{ssec: choice of parameters}.}
    \label{tab: explicit bounds}
\end{table}
                %\gr{Move the table here, add comment from the appendix here about sometimes one bound is better, sometimes the other}

\section{Conclusion and Future Works}\label{sec:con}

In this work, we proposed an algorithm that jointly generates a network representation and a private synthetic network from a given dataset. We  provided a theoretical analysis of our algorithm studying the accuracy w.r.t. the fused Gromov Wasserstein distance and comparing the corresponding network distributions. Under additional assumptions, we derived optimal parameter choices and deduced that in this case our bounds are $\mathcal{O}(n^{-1/(d+1)})$.

An interesting further direction would be the consideration of directed networks. %This would require a thorough adaptation of our graph model. 
Furthermore, it would be worthwhile to study networks with weighted edges. Here challenges lie in a possible dependence of the weight on the existence of an edge. Moreover, it would require a new notion of accuracy for attributed networks with weighted edges. %This not only raises the question of whether our proofs can be adapted to such a dependency, but also requires a careful analysis of the privacy. 

It would be interesting to combine the node privacy studied in this paper with a suitable notion of edge privacy. This could be achieved, for example, by adapting the rule how edges are being created by including a resampling mechanism on the edges. However, a suitable notion of such a joint privacy measure to our knowledge has not yet been investigated. 

Finally, we intend to extend our results to include $(\epsilon, \delta)$- differential privacy, see \citet{dwork2006differential}, thus allowing for more discrete noise distributions, e.g., a  discrete Gaussian distribution~\citep{canonne2020discrete}.
%, by considering $(\epsilon,\delta)$-differential private definition, \citep{dwork2006differential}. 

 %\newpage
%\clearpage
% 
\bibliographystyle{plainnat}
\bibliography{references}

\newpage
% \clearpage
 \appendix
\section{More Related Works}
This section details related works about differential privacy in graphs.

\textbf{Differential Privacy in Graphs:} Vertex and edge differential privacy in graphs were introduced in \citep{kasiviswanathan2013analyzing} using the rewiring distance. Several works focus on differentially private graph publishing~\citep{jian2021publishing,day2016publishing,hou2023ppdu}. Other works propose practical approaches for private synthetic graph generation under different notions of privacy~\citep{qin2017generating,zahirnia2024neural}. In contrast to these approaches, our work provides a theoretical foundation for private synthetic graph generation, with rigorous guarantees.

 \section{Proofs and details of Section~\ref{sec:prem}}\label{app:sec_prem}
% \begin{tcolorbox}
%      \begin{repproposition}{prop:DP-gaurantee}
%     Assuming $\lambda_i\sim\mathcal{N}_{\mathbb{Z}}(0,n^2/\epsilon^2)$, Algorithm~\ref{alg: PSMM} is $\epsilon$-differential privacy at  vertex-level.
% \end{repproposition}
% \end{tcolorbox}

% \begin{proof}
%     The proof follows directly from combining \citep[Theorem~4]{canonne2020discrete} with \citep[Proposition~3.4]{he2023}. Note that $\frac{1}{2\epsilon^2}$-concentrated differential privacy is equal to $\epsilon$-differential privacy.
% \end{proof}
% {\color{red}{GR: I am not quite following the proof. Canonne et al. Th 4 is for discrete Gaussian, not Gaussian. Also $\epsilon$-DP implies $1/2 \epsilon^2$-concentrated DP but not the other way round, they are not equivalent. In \citet{he2023} it should probably be Theorem 1 that we refer to, but that is for Laplacian noise.}}\todoa{You are right. We can just prove $(\epsilon,\delta)$-differential privacy under discrete Gaussian noise.}

 \begin{tcolorbox}
\begin{repproposition}{prop:total_measure}
Algorithm~\ref{alg: linear_programming} outputs the closest probability measure $\hnu$ on $\mcy$ to a given discrete signed measure~$\nu$ on $\mcy$ with respect to the total variation distance.
\end{repproposition}
 \end{tcolorbox}
\begin{proof}
Let $\nu$ be a discrete signed probability measure on $\mcy$. Then for any probability measure $\tau$ on $\mcy$ the total variation distance between $\nu$ and $\tau$ is given by
$$d_{TV}(\nu,\tau) = \sum_{i=1}^m \abs{\nu_i - \tau_i},$$
where $\nu_i := \nu(\{y_i\})$ and $\tau_i := \tau(\{y_i\})$, see Section~\ref{ssec: TV}. Thus, finding the closest probability measure with respect to the total variation distance is equivalent to solving the program~\eqref{eq: P Algorithm2} given by
\begin{align} \label{eq: P Algorithm2}
        \min \quad & \sum_{i=1}^m \abs{\nu_i - \tau_i} \tag{P$1$}\\
        \mbox{s.t.}\quad & \sum_{i=1}^m \tau_i = 1, \quad \tau_i \geq 0, \quad \forall i\leq m.\notag
\end{align}
Then as $\abs{\nu_i - \tau_i} = \max\{\nu_i - \tau_i,  \tau_i -\nu_i\}$, minimizing the absolute value~$\abs{\nu_i - \tau_i}$ is equivalent to finding the minimal $u_i\geq \max\{\nu_i - \tau_i,  \tau_i -\nu_i\}$. In particular, \eqref{eq: P Algorithm2} is equivalent to solving the linear program in Algorithm~\ref{alg: linear_programming}.
\end{proof} 
% \newpage
\section{Proofs and details of Section~\ref{sec:Main}}\label{app:sec:main}
\begin{tcolorbox}
\begin{reptheorem}{thm: Algorithm 3 distribution}
The graphs $(\Xi,\Sigma)$ and $(\Eta,\Tau)$ obtained by Algorithm~\ref{alg: graph generation} have distributions $P^{\mu^b_{\mcy}}\otimes Q^{\kappa}$ and $P^{\mu^a_{\mcx}}\otimes Q^{\kappa}$, respectively.
\end{reptheorem}
\end{tcolorbox}
\begin{proof}
   We show that the two attributed random graphs constructed in Algorithm~\ref{alg: graph generation} define a coupling of the graph models introduced in Section~\ref{ssec: graph model}. We write $\Prob_{\lambda}$ and $\E_{\lambda}$ to denote the conditional probability and expectation given $\lambda = (\lambda_i)_{i=1}^m$. Recall that the probability measure $\hnu$ depends on $\lambda$. We write $(\Eta,\Tau)$ and $(\Xi,\Sigma)$ for the graphs constructed according to the graph model in Section~\ref{ssec: graph model} and $(\hat{\Eta},\hat{\Tau})$ as well as $(\hat{\Xi},\hat{\Sigma})$ for the graphs constructed in Algorithm~\ref{alg: graph generation}.

\textit{Step 1: Coupling of the vertex process}\\
Consider the total number of vertices $N\sim \text{Poi}(a)$ and $M\sim \text{Poi}(b)$ and assume without loss of generality that $a\geq b$. Then $N \overset{d}{=} M + K_N$ for $K_N\sim \text{Poi}(a-b)$. In particular, this defines a coupling of $N$ and $M$ such that $N\geq M$.

For $k\in[m]$, let $N_k := \Eta(\Omega_k)$ be the number of points~$(X_i,U_i)$ of $\Eta$ whose attributes $X_i$ are in $\Omega_k$. As the $X_i$ were chosen uniformly from $\mcx$ we obtain 
$$\Prob(X_i\in \Omega_k) = \frac{\#\{x_j\in \Omega_k: j\in[n]\}}{n} = \frac{n_k}{n}.$$
Given the total number of points~$N$ in $\eta$, the  vector $\mathbb{N}:=(N_1,\ldots,N_m)$  of attribute counts in the partition sets $(\Omega_i)_{i\in[m]}$ has a multinomial distribution,
%with these parameters, i.e. 
$\mathbb{N}\mvert N\sim \text{Multin}(N,(\frac{n_1}{n},\ldots,\frac{n_m}{n})).$ Analogously, for $k\in[m]$ let $M_k := \Xi(\Omega_k)$ be the number of points $(Y_i,V_i)$ of $\Xi$ whose attributes $Y_i$ are in $\Omega_k$. Then we have 
$$\Prob_{\lambda}(Y_i\in \Omega_k) = \Prob(Y_i = y_k) = \hat{\nu}_k$$
and, given the total number of points~$M$ in $\Xi$, we obtain that the conditional distribution is $\mathbb{M}\mvert M:=(M_1,\ldots,M_m)\mvert M\sim \text{Multin}(M,(\hat{\nu}_1,\ldots,\hat{\nu}_m))$. 

We show that the vertex counts constructed in Algorithm~\ref{alg: graph generation} obtain the same distribution.
%We start with the construction of a coupling of these attribute counts $\N$ and $\mathbb{M}$. 
Let $\mathbb{Z}^j:=(Z^j_1,\ldots,Z^j_m)\in\{0,1\}^m$, $j\in [M]$, be independent random variables defined by
\begin{align*}
    \Prob_{\lambda}(\mathbb{Z}^j = (l_1,\ldots,l_m)) &=\prod_{k\in[m]}\bigg(\frac{n_k}{n}\wedge \hat{\nu}_k\bigg)^{l_k}, \quad \mbox{ for } l_1,\ldots,l_m\in\{0,1\} \mbox{ such that } \sum_{k\in[m]} l_k = 1; \\
%\end{align*}
%for $l_1,\ldots,l_m\in\{0,1\}$ such that $\sum_{k\in[m]} l_k = 1$ and 
%\begin{align*}
    \Prob_{\lambda}(\mathbb{Z}^j = (0,\ldots,0)) &= 1- \sum_{\substack{l_1,\ldots,l_m\in\{0,1\}\\\sum_{k\in[m]} l_k = 1}} \Prob_{\lambda}(\mathbb{Z}^j = (l_1,\ldots,l_m)) =1- \sum_{k\in[m]} \frac{n_k}{n}\wedge \hat{\nu}_k.
\end{align*}
Note that this construction corresponds to the choice of common vertex counts in Algorithm~\ref{alg: graph generation}. 

We describe the construction of non-common vertex counts. First, define $\mathbb{K}\mvert (N-M) := (K_1,\cdots,K_m)\mvert (N-M) \sim \text{Multin}(N-M,(\frac{n_1}{n},\ldots,\frac{n_m}{n}))$.

If $\frac{n_k}{n} = \hat{\nu}_k$ for all $k\in[m]$, then $\mathbb{N} \overset{d}{=}\mathbb{M} + \mathbb{K}$ and we obtain a coupling $\N$ and $\mathbb{M}$ such that $ N_k = M_k + K_k = \sum_{j\in [M]} Z_k^j + K_k$ for all $k\in [m]$. 

Otherwise, there exists a $k\in[m]$ such that $\frac{n_k}{n} \neq \hat{\nu}_k$ and we obtain $\Prob_{\lambda}(\mathbb{Z}^j = (0,\ldots,0))>0$. For $j\in[M]$ we then define $\mathbb{Z}^{N,j} := (Z_1^{N,j},\ldots,Z_m^{N,j})$ by 
\begin{align*}
\Prob_{\lambda}(\mathbb{Z}^{N,j} = (l_1,\ldots,l_m)) = \big(\Prob_{\lambda}(\mathbb{N}^j = (l_1,\ldots,l_m)) - \Prob_{\lambda}(\mathbb{Z}^j = (l_1,\ldots,l_m))\big)\Prob_{\lambda}(\mathbb{Z}^j = (0,\ldots,0))^{-1},
\end{align*}
where $\mathbb{N}^j \sim \text{Multin}(1,(\frac{n_1}{n},\ldots,\frac{n_m}{n}))$ and  $l_1,\ldots,l_m\in\{0,1\}$ such that $\sum_{k\in[m]} l_k = 1$. Note that this defines a probability distribution as  
\begin{align*}
&\sum_{\substack{l_1,\ldots,l_m\in\{0,1\}\\\sum_{k\in[m]} l_k = 1}}\Prob_{\lambda}(\mathbb{Z}^{N,j} = (l_1,\ldots,l_m))
%&=\sum_{\substack{l_1,\ldots,l_m\in[N]\\\sum_{k\in[m]} l_k = 1}}\big(\Prob_{\lambda}(\mathbb{N}^j = (l_1,\ldots,l_m)) - \Prob_{\lambda}(\mathbb{Z}^j = (l_1,\ldots,l_m))\big)\Prob_{\lambda}(\mathbb{Z}^j = (0,\ldots,0))^{-1}
\\
&= \bigg(1 - \sum_{\substack{l_1,\ldots,l_m\in\{0,1\}\\\sum_{k\in[m]} l_k = 1}}\Prob_{\lambda}(\mathbb{Z}^j = (l_1,\ldots,l_m))\bigg)\Prob_{\lambda}(\mathbb{Z}^j = (0,\ldots,0))^{-1} =1.
\end{align*}
Furthermore, $\mathbb{Z}^{N,j} = (l_1,\ldots,l_m)\mvert (\Z^j = (0,\ldots,0))\sim \text{Multin}(1,(\frac{n_1}{n},\ldots,\frac{n_m}{n}))$ and thus $(Z_1^{N,j} + K_1,\ldots,Z_m^{N,j} + K_m)\sim \text{Multin}(N-Z,(\frac{n_1}{n},\ldots,\frac{n_m}{n}))$ for $Z = \sum_{j\in [M]}\sum_{k\in [m]} Z_k^j$. In particular, the construction of $\mathbb{Z}^{N,j}$ and $\mathbb{K}$ corresponds to the creation of non-common vertex counts in Algorithm~\ref{alg: graph generation}.
Analogously we define $\mathbb{Z}^{M,j}$ and set
$\hat{\mathbb{N}} = \sum_{j\in[M]} \hat{\mathbb{N}}^j + \sum_{j\in[N]\setminus[M]} {\mathbb{N}}^j = \sum_{j\in[M]}\mathbb{Z}^j +\mathbb{Z}^{N,j}\1_{\{\mathbb{Z}^j = (0,\ldots,0)\}}+ \mathbb{L}$ as well as $\hat{\mathbb{M}} = \sum_{j\in[M]} \hat{\mathbb{M}}^j = \sum_{j\in[M]}\mathbb{Z}^j +\mathbb{Z}^{M,j}\1_{\{\mathbb{Z}^j = (0,\ldots,0)\}}$. Then $(\hat{\mathbb{N}},\hat{\mathbb{M}})$ is a coupling of $\mathbb{N}$ and ${\mathbb{M}}$ as for any $j\in [M]$ 
\begin{align*}
    &\Prob_{\lambda}(\hat{\mathbb{N}}^j =(l_1,\ldots,l_m)) \\
    &=\Prob_{\lambda}(\hat{\mathbb{N}}^j =(l_1,\ldots,l_m),\mathbb{Z}^j = (0,\ldots,0) ) +\Prob_{\lambda}(\hat{\mathbb{N}}^j =(l_1,\ldots,l_m),\mathbb{Z}^j \neq (0,\ldots,0) )\\
    &= \Prob_{\lambda}(\mathbb{Z}^{N,j} =(l_1,\ldots,l_m)) \Prob_{\lambda}(\mathbb{Z}^j = (0,\ldots,0) ) +\Prob_{\lambda}({\mathbb{Z}^j} =(l_1,\ldots,l_m) )\\
    &= \Prob_{\lambda}({\mathbb{N}}^j =(l_1,\ldots,l_m)) 
\end{align*}
and analogously $\Prob_{\lambda}(\hat{\mathbb{M}}^j =(l_1,\ldots,l_m))= \Prob_{\lambda}({\mathbb{M}}^j =(l_1,\ldots,l_m))$.
Furthermore, by the construction of $\mathbb{Z}$ we obtain for any $j\in[M]$ that $$\Prob_{\lambda}(\hat{\mathbb{N}}^j\neq \hat{\mathbb{M}}^j) = \Prob_{\lambda}(\mathbb{Z}^j = (0,\ldots,0)).$$

We now use the fact that the vertex counts in Algorithm~\ref{alg: graph generation} have the correct marginal distributions to follow that the vertex processes $\hat{\Eta}$ and $\hat{\Xi}$ have the correct marginal distributions. Recall that we constructed $\hat{\Eta} = \sum_{i=1}^m\sum_{j=1}^{\hat{N_i}} \delta_{(X_{ij},U_{ij})}$ for $X_{ij}\sim \mathcal{U}(\mcx\cap \Omega_i)$, $U_{ij}\sim\mcu([0,1])$ and $\hat{\Xi} = \sum_{i=1}^m\sum_{j=1}^{\hat{M_i}} \delta_{(Y_{ij},U_{ij})}$ for $Y_{ij}=y_i$. Note that the distributions of the attributes, the identifiers, and the number of attributes in each partition set are unchanged compared to the graph model in Section~\ref{ssec: graph model}. Hence, this defines a coupling $(\hat{\Eta},\hat{\Xi})$ of the vertex processes $\Eta$ and $\Xi$.

\textit{Step 2: Coupling of the edge process}\\
Based on the coupling of the vertex processes we can now show that the edge processes $(\hat{\Tau},\hat{\Sigma})$ constructed in Algorithm~\ref{alg: graph generation} define a coupling of the edge processes $\Tau$ and $\Sigma$. Recall that the existence of an edge~$(U_{ij},U_{i'j'})$ in $\Tau$ and $\Sigma$ is determined by Bernoulli random variables $K_{ij}^{\Tau}$ and $K_{ij}^{\Sigma}$, respectively, see Section~\ref{ssec: graph model}. Furthermore, the connection probabilities in Algorithm~\ref{alg: graph generation} were defined by 
\begin{align*}
\Prob_{\lambda}((\hat{K}^{\Tau}_{iji'j'},\hat{K}^{\Sigma}_{iji'j'})=(1,1))&=\kappa(X_{ij},X_{i'j'})\wedge \kappa(Y_{ij},Y_{i'j'})\\
\Prob_{\lambda}((\hat{K}^{\Tau}_{iji'j'},\hat{K}^{\Sigma}_{iji'j'})=(1,0))&=\kappa(X_{ij},X_{i'j'})-(\kappa(X_{ij},X_{i'j'})\wedge \kappa(Y_{ij},Y_{i'j'}))\\  
\Prob_{\lambda}((\hat{K}^{\Tau}_{iji'j'},\hat{K}^{\Sigma}_{iji'j'})=(0,1))&=\kappa(Y_{ij},Y_{i'j'})-(\kappa(X_{ij},X_{i'j'})\wedge \kappa(Y_{ij},Y_{i'j'}))\\
\Prob_{\lambda}((\hat{K}^{\Tau}_{iji'j'},\hat{K}^{\Sigma}_{iji'j'})=(0,0))&=1-\kappa(X_{ij},X_{i'j'})\vee \kappa(Y_{ij},Y_{i'j'})
\end{align*}
for any $i,i'\in[m]$ and any $j\in[Z_i]$ and $j'\in [Z_{i'}]$ and by $\hat{K}^{\Tau}_{iji'j'}\sim \text{Ber}(\kappa(X_{ij},X_{i'j'}))$ and $\hat{K}^{\Sigma}_{iji'j'}\sim \text{Ber}(\kappa(Y_{ij},Y_{i'j'}))$ for any $(j,j')$ such that $j>Z_i$ or $j'>Z_{j'}$. 
Clearly, this construction of $\hat{K}^{\Tau}$ and $\hat{K}^{\Sigma}$ corresponds to a maximal coupling of the two Bernoulli random variables $K_{ij}^{\Tau}$ and $K_{ij}^{\Sigma}$. In particular, 
$$\Prob_{\lambda}(\hat{K}^{\Tau}_{iji'j'} \neq \hat{K}^{\Sigma}_{iji'j'}) %= 1-\big(\kappa(X_{ij},X_{i'j'})\wedge \kappa(Y_{ij},Y_{i'j'}) + 1 - \kappa(X_{ij},X_{i'j'})\vee \kappa(Y_{ij},Y_{i'j'})\big) 
%= \kappa(X_{ij},X_{i'j'})\vee \kappa(Y_{ij},Y_{i'j'}) - \kappa(X_{ij},X_{i'j'})\wedge \kappa(Y_{ij},Y_{i'j'})
= \abs{\kappa(X_{ij},X_{i'j'})- \kappa(Y_{ij},Y_{i'j'})}.$$
Thus, setting 
$$\hat{\Tau} = \sum_{i,i'=1}^m\sum_{j=1}^{\hat{N}_i}\sum_{j'=1}^{\hat{N}_{i'}} \hat{K}^{\Tau}_{iji'j'} \delta_{\{U_{ij},U_{i'j'}\}} %$$
%and
\mbox{ and } 
%$$
\hat{\Sigma} = \sum_{i,i'=1}^m\sum_{j=1}^{\hat{M}_i}\sum_{j'=1}^{\hat{M}_{i'}} \hat{K}^{\Sigma}_{iji'j'} \delta_{\{U_{ij},U_{i'j'}\}} $$
yields a coupling of the edge processes. Overall, this shows that $(\hat{\Eta},\hat{\Tau})$ and $(\hat{\Xi},\hat{\Sigma})$ constructed in Algorithm~\ref{alg: graph generation} yield a coupling of $({\Eta},{\Tau})$ and $({\Xi},{\Sigma})$ constructed in Section\ref{ssec: graph model} and thus have the claimed (marginal) distributions.
\end{proof}
\begin{tcolorbox}
\begin{reptheorem}{thm: result coupling}
Let $(\Eta,\Tau)$ and $(\Xi,\Sigma)$ be the ``true'' and private synthetic random graphs of expected size $a>0$ and $b>0$, respectively, constructed by Algorithm~\ref{alg: graph generation}. % using a probability measure $\hnu$ created by Algorithm~\ref{alg: PSMM} with random noise $\Lambda$. 
Then 
\begin{align*}
    \E\big(d_{FGW}(\eta^{(\Eta,\Tau)},\eta^{(\Xi,\Sigma)})\big) %\leq \tilde{C}_1 \big(\max_{k\in[m]}\mathrm{diam}(\Omega_k)\big)  \sum_{k\in[m]}  \frac{n_k}{n}\wedge \hat{\nu}_k+2 \tilde{C}_2\frac{m}{n} \E\abs{\Lambda}
    &\leq \tilde{C}_1\,\max_{k\in[m]}\mathrm{diam}(\Omega_k) \frac{1}{1+\frac{a\wedge b}{\abs{a-b}}\1_{\{\abs{a-b}>0\}}} + 4\tilde{C}_2 \frac{m}{n} \E(\abs{\Lambda})\\
    &+ \tilde{C}_2 \bigg(1 - \frac{1}{\big(1+\frac{a\wedge b}{\abs{a-b}}\big)^2}\bigg) \1_{\{\abs{a-b}>0\}},
\end{align*}
where $\tilde{C}_1 = 1-\alpha+\alpha (2\,C\,L_{\kappa}) $ and $\tilde{C}_2 =  (1-\alpha)\mathrm{diam}(\Omega)+\alpha \min\{C,2\,C\,L_{\kappa} \mathrm{diam}(\Omega)\} $ with $\alpha\in [0,1]$ and $C>0$ parameters of the FGW metric and $L_{\kappa}$ the Lipschitz constant of edge connection function~$\kappa$.
\end{reptheorem}
\end{tcolorbox}
% \subsection{Proof of Theorem~\ref{thm: result coupling}}
\begin{proof}
We write $\Prob_{\lambda}$ and $\E_{\lambda}$ to denote the conditional probability and expectation given $\lambda = (\lambda_i)_{i=1}^m$ and assume w.l.o.g. that $a\geq b$. We stick to the notation introduced in the Proof of Theorem~\ref{thm: Algorithm 3 distribution}, see App. \ref{app:sec:main}. In particular, we denote the graphs constructed with Algorithm~\ref{alg: graph generation} by $(\hat{\Eta},\hat{\Tau})$ and $(\hat{\Xi},\hat{\Sigma})$ with 
$$\hat{\Eta} = \sum_{i=1}^m\sum_{j=1}^{\hat{N_i}} \delta_{(X_{ij},U_{ij})}\quad \text{ and }\quad\hat{\Xi} = \sum_{i=1}^m\sum_{j=1}^{\hat{M_i}} \delta_{(Y_{ij},U_{ij})}$$ as well as 
$$\hat{\Tau} = \sum_{i,i'=1}^m\sum_{j=1}^{\hat{N}_i}\sum_{j'=1}^{\hat{N}_{i'}} \hat{K}^{\Tau}_{iji'j'} \delta_{\{U_{ij},U_{i'j'}\}} \quad \text{ and }\quad\hat{\Sigma} = \sum_{i,i'=1}^m\sum_{j=1}^{\hat{M}_i}\sum_{j'=1}^{\hat{M}_{i'}} \hat{K}^{\Sigma}_{iji'j'} \delta_{\{U_{ij},U_{i'j'}\}}.$$

\textit{Step 1: Derivation of the upper bound}\\
We use the joint construction of the vertex- and edge processes to calculate the expected distance of the two graphs with respect to the FGW metric, see Section~\ref{ssec: FGW}. 
Recall that we coupled $\abs{\hat{\Eta}} = N$ and $\abs{\hat{\Xi}} = M$ such that $N = M + K$ for $K\sim \text{Poi}(a-b)$. By conditioning on the (minimal) number of vertices $(X_j,Y_j)$ that are in the same cell, we obtain 
\begin{align}\label{eq: coupling proof}
    &\E_{\lambda}(d_{FGW}(\eta^{(\Eta,\Tau)},\eta^{(\Xi,\Sigma)}))\notag\\
    &= \E_{\lambda}\big[\E_{\lambda}(d_{FGW}(\eta^{(\hat{\Eta},\hat{\Tau})},\eta^{(\hat{\Xi},\hat{\Sigma})})\mvert N,M)\big]\nonumber  \\
    &= \E_{\lambda}\bigg[\sum_{l=0}^{M}\E_{\lambda}(d_{FGW}(\eta^{(\hat{\Eta},\hat{\Tau})},\eta^{(\hat{\Xi},\hat{\Sigma})})\,\bigg\vert\, 
    \sum_{j\in [M]} \norm{\Z^j} = l, %\mathbb{Z}^j \neq (0,\ldots,0) \text{ for $l$ many $j$}, 
    N,M)\,\\
    &\qquad\binom{M}{l} \Prob_{\lambda}( \mathbb{Z}^1\neq (0,\ldots,0)) ^l\, \Prob_{\lambda}( \mathbb{Z}^1=  (0,\ldots,0))^{(M-l)} \bigg]\notag.
\end{align}
%\gr{What do we mean by $\mathbb{Z}^j \neq (0,\ldots,0) \text{ $l$ times}$?}\leoni{I rephrased it.}
Let $(X_{ij},U_{ij}),(X_{i'j'},U_{i'j'})\in \hat{\Eta}$ and $(Y_{kl},U_{kl}),(Y_{k'l'},U_{k'l'})\in\hat{\Xi}$. Then 
$$d((X_{ij},U_{ij}),(Y_{kl},U_{kl})) := d(X_{ij},Y_{kl})\leq \mathrm{diam}(\Omega)$$ and 
\begin{align*}
    \abs{d_{\hat{\Tau}}(U_{ij},U_{i'j'}) - d_{\hat{\Sigma}}(U_{kl},U_{k'l'})} \leq C \, \Prob(\hat{K}^{\Tau}_{iji'j'} \neq \hat{K}^{\Sigma}_{klk'l'}) &= \min\{C,C\, \abs{\kappa(X_{ij},X_{i'j'})-\kappa(Y_{kl},Y_{k'l'})} \}\\ &\leq \min\{C,2\,C\,L_{\kappa} \mathrm{diam}(\Omega)\},
\end{align*} 
where the last inequality follows as 
 the Lipschitz continuity of $\kappa$ implies
\begin{align}\label{eq: lipschitz argument kappa}
    \abs{\kappa(x,x')-\kappa(y,y')} &\leq \abs{\kappa(x,x')-\kappa(y,y')} + \abs{\kappa(y,x')-\kappa(y,y')} \\
    &\leq L_{\kappa} \,(d(x,y) + d(x',y')). \nonumber 
\end{align}

If $i=k$ and $i'=k'$ we have by construction $X_{ij},Y_{il}\in\Omega_i$ and $X'_{i'j'},Y_{i'l'}\in\Omega_{i'}$. Hence, we can improve the above bounds to $$d(X_{ij},Y_{il})\leq \mathrm{diam}(\Omega_i) \leq \max_{k\in[m]}\mathrm{diam}(\Omega_k) $$ and 
\begin{align*}
    \abs{d_{\hat{\Tau}}(U_{ij},U_{i'j'}) - d_{\hat{\Sigma}}(U_{il},U_{i'l'})} &\leq C \, \Prob(\hat{K}^{\Tau}_{iji'j'} \neq \hat{K}^{\Sigma}_{ili'l'}) 
    %= C\, \abs{\kappa(X_{ij},X_{i'j'})-\kappa(Y_{il},Y_{i'l'})}   
    \leq 2\,C\,L_{\kappa} \max_{k\in[m]}\mathrm{diam}(\Omega_k).
\end{align*} 
%\leoni{I commented something out here. It is already stated in equation \ref{eq: coupling proof} above and in the text below.}
% By conditioning on the (minimal) number of vertices $(X_j,Y_j)$ that are in the same cell, we obtain
% \begin{align}\label{eq: coupling proof2}
%     &\E(d_{FGW}(\eta^{(\Eta,\Tau)},\eta^{(\Xi,\Sigma)}))\notag\\
%     &= \E\big[\E(d_{FGW}(\eta^{(\hat{\Eta},\hat{\Tau})},\eta^{(\hat{\Xi},\hat{\Sigma})})\mvert N,M)\big]\nonumber  \\
%    &= \E\bigg[\sum_{l=0}^{N\wedge M}\E(d_{FGW}(\eta^{(\hat{\Eta},\hat{\Tau})},\eta^{(\hat{\Xi},\hat{\Sigma})})\mvert \mathbb{Z}^j \neq (0,\ldots,0) \text{  for $l$ many $j$}, N,M)\,\\
%     &\qquad\binom{N\wedge M}{l} \Prob( \mathbb{Z}^1\neq (0,\ldots,0)) ^l\, \Prob( \mathbb{Z}^1=  (0,\ldots,0))^{(N\wedge M-l)} \bigg]\notag.
% \end{align}
%Then as $\mathbb{Z}^j \neq (0,\ldots,0)$  for $l$ many $j$, we have at least $l$ %many 
%points of $\hat{\Eta}$ and $\hat{\Xi}$ that have attributes in the same cell, and $N-l$ and $M-l$ %many 
%points, respectively, that have attributes in different partition sets. 
Now consider the conditional expectation in Equation~\eqref{eq: coupling proof}. 
As $ \sum_{j\in [M]} \norm{\Z^j} = l$, i.e. $\mathbb{Z}^j \neq (0,\ldots,0)$ for $l$ many $\mathbb{Z}^j$, we have that $\sum_{k\in[m]}Z_k = \sum_{k\in[m]}\sum_{j\in [M} Z_k^j = l$. In particular, we have at least $l$ many points of $\hat{\Eta}$ and $\hat{\Xi}$ that have attributes in the same partition set and $N-l$ and $M-l$ many points, respectively, that have attributes in different partition sets. We calculate an upper bound of the FGW distance by restricting to transport plans that move mass $1/N$ from vertex $X_{ij}$ (having mass $1/N$) to vertex $Y_{ij}$ (having mass $1/M$) for all $i\in[m]$ and $j\in [Z_i]$ and distribute the remaining mass $(1-l/N)$ according to some subcoupling of the remaining vertices. Then
\begin{align*}
    &\E_{\lambda}\bigg(d_{FGW}(\eta^{(\hat{\Eta},\hat{\Tau})},\eta^{(\hat{\Xi},\hat{\Sigma})})\,\bigg\vert\, \sum_{j\in [M]} \norm{\Z^j} = l, N,M\bigg)\\
    &\leq \sum_{i,i'=1}^m\sum_{j=1}^{Z_i}\sum_{j=1}^{Z_{i'}} \E_{\lambda}\big[ \big((1-\alpha) d(X_{ij},Y_{ij}) + \alpha \abs{d_{\hat{\Tau}}(U_{ij},U_{i'j'}) - d_{\hat{\Sigma}}(U_{ij},U_{i'j'})}\big) \big]\frac{1}{N^2} \\
    &\hspace{0.2cm}+ \frac{N^2-l^2}{N^2} \big((1-\alpha)\mathrm{diam}(\Omega)+\alpha \min\{C,2\,C\,L_{\kappa} \mathrm{diam}(\Omega)\}\big)\\
    &\leq \frac{l^2}{N^2} \big(1-\alpha+\alpha (2\,C\,L_{\kappa})\big) \big(\max_{k\in[m]}\mathrm{diam}(\Omega_k)\big)\\
    &+ \frac{N^2-l^2}{N^2} \big((1-\alpha)\mathrm{diam}(\Omega)+\alpha \min\{C,2\,C\,L_{\kappa} \mathrm{diam}(\Omega)\}\big) ,   
\end{align*}
where we used the above bounds, applying the special case to the first term and the general case to the second term.

Recall that by the construction of $\mathbb{Z}^1$ we have
$\Prob_{\lambda}(\mathbb{Z}^1 \neq (0,\ldots,0)) = \sum_{k\in[m]}  \Big( \frac{n_k}{n}\wedge \hat{\nu}_k \Big).$ We set $\tilde{C}_1 := \big(1-\alpha+\alpha (2\,C\,L_{\kappa})\big) $ and  $\tilde{C}_2 :=(1-\alpha)\mathrm{diam}(\Omega)+\alpha \min\{C,2\,C\,L_{\kappa} \mathrm{diam}(\Omega)\} .$ We let $L\mvert {M}\sim \text{Bin}({M},\Prob_{\lambda}( \mathbb{Z}^1\neq (0,\ldots,0)))$. Taking the expectation of $\lambda$ and using Equation~\eqref{eq: coupling proof} yields 
\begin{align*}
    &\E(d_{FGW}(\eta^{(\hat{\Eta},\hat{\Tau})},\eta^{(\hat{\Xi},\hat{\Sigma})})) =\E\bigg[\E_{\lambda}(d_{FGW}(\eta^{(\hat{\Eta},\hat{\Tau})},\eta^{(\hat{\Xi},\hat{\Sigma})}))\,  \bigg]  \\
    &= \E\bigg[\E_{\lambda}\bigg[\sum_{l=0}^{M}\E_{\lambda}(d_{FGW}(\eta^{(\hat{\Eta},\hat{\Tau})},\eta^{(\hat{\Xi},\hat{\Sigma})})\mvert \sum_{j\in [M]} \norm{\Z^j} = l, N,M)\,\\
    &\hspace{4cm}\binom{{M}}{l} \Prob_{\lambda}( \mathbb{Z}^1\neq (0,\ldots,0)) ^l\, \Prob_{\lambda}( \mathbb{Z}^1=  (0,\ldots,0))^{({M}-l)}\bigg]\,  \bigg]\\
    &\leq \E\bigg[\E_{\lambda}\bigg[\sum_{l=0}^{M}\binom{{M}}{l} \bigg(\frac{l^2}{{N}^2} \tilde{C}_1 \big(\max_{k\in[m]}\mathrm{diam}(\Omega_k)\big)  + \frac{{N}^2-l^2}{{N}^2} \tilde{C}_2\bigg) \, \\
    &\hspace{4cm}\Prob_{\lambda}( \mathbb{Z}^1\neq (0,\ldots,0)) ^l\, \Prob_{\lambda}( \mathbb{Z}^1=  (0,\ldots,0))^{({N}-l)}\bigg]\,  \bigg]\\ 
    %\bigg(\sum_{k\in[m]}  \frac{n_k}{n}\wedge \hat{\nu}_k\bigg) ^l\, \bigg(1-\sum_{k\in[m]}  \frac{n_k}{n}\wedge \hat{\nu}_k\bigg)^{({N}-l)}\\
    &= \E\bigg[\E_{\lambda}\bigg[\tilde{C}_1\big(\max_{k\in[m]}\mathrm{diam}(\Omega_k)\big)  \frac{\E_{\lambda}(L^2\mvert M)}{{N}^2} + \tilde{C}_2\bigg(1-\frac{\E_{\lambda}(L^2\mvert M)}{{N}^2}\bigg)\bigg]\,  \bigg]\\
    &= \E\bigg[\E_{\lambda}\bigg[\tilde{C}_1\big(\max_{k\in[m]}\mathrm{diam}(\Omega_k)\big)  \bigg[\frac{M}{{N^2}}\Prob_{\lambda}( \mathbb{Z}^1\neq (0,\ldots,0)) + \frac{M^2-M}{{N^2}} \Prob_{\lambda}( \mathbb{Z}^1\neq (0,\ldots,0))^2\bigg]\\
    &\hspace{0.3cm}+ \tilde{C}_2\bigg[1 - \frac{M}{{N^2}}\Prob_{\lambda}( \mathbb{Z}^1\neq (0,\ldots,0)) - \frac{M^2-M}{{N^2}} \Prob_{\lambda}( \mathbb{Z}^1\neq (0,\ldots,0))^2\bigg]\bigg]\,  \bigg]\\
    &\leq \E\bigg[ \E_{\lambda}\bigg[\tilde{C}_1\big(\max_{k\in[m]}\mathrm{diam}(\Omega_k)\big)  \bigg[\frac{M}{{N^2}}\Prob_{\lambda}( \mathbb{Z}^1\neq (0,\ldots,0)) + \frac{M^2-M}{{N^2}} \Prob_{\lambda}( \mathbb{Z}^1\neq (0,\ldots,0))\bigg]\\
    &\hspace{0.3cm}+ \tilde{C}_2\bigg[\frac{M}{{N^2}}\bigg(1 -\Prob_{\lambda}( \mathbb{Z}^1\neq (0,\ldots,0))\bigg) + \frac{N^2-M}{{N^2}} \bigg(1^2 -\Prob_{\lambda}( \mathbb{Z}^1\neq (0,\ldots,0))^2\bigg)\\
    &\hspace*{3cm}- \frac{M^2-N^2}{{N^2}} \Prob_{\lambda}( \mathbb{Z}^1\neq (0,\ldots,0))^2\bigg]\bigg]\,  \bigg]\\
    &\leq \E\bigg[\E_{\lambda}\bigg[\tilde{C}_1\big(\max_{k\in[m]}\mathrm{diam}(\Omega_k)\big)  \frac{M^2}{{N}^2} \sum_{k\in[m]}  \Big( \frac{n_k}{n}\wedge \hat{\nu}_k \Big) \\
    &\hspace{0.3cm}+ \tilde{C}_2\bigg[\frac{M}{{N^2}}\bigg(1 -\sum_{k\in[m]}  \Big( \frac{n_k}{n}\wedge \hat{\nu} _k \Big) \bigg) + 2\bigg(1-\frac{M}{{N^2}}\bigg) \bigg(1 -\sum_{k\in[m]} \Big(  \frac{n_k}{n}\wedge \hat{\nu}_k \Big) \bigg)\\
    &\hspace*{3cm}+ \bigg(1-\frac{M^2}{{N^2}}\bigg) \bigg(\sum_{k\in[m]}  \Big( \frac{n_k}{n}\wedge \hat{\nu}_k \Big) \bigg)^2\bigg]\bigg]\,  \bigg]\\
    %&\leq \tilde{C}_1\big(\max_{k\in[m]}\mathrm{diam}(\Omega_k)\big)  \sum_{k\in[m]}  \frac{n_k}{n}\wedge \hat{\nu}_k + \tilde{C}_2\bigg(2-\frac{1}{{N}}\bigg) \bigg(1 -\sum_{k\in[m]}  \frac{n_k}{n}\wedge \hat{\nu}_k\bigg)\\
    &\leq \E\bigg[\tilde{C}_1\big(\max_{k\in[m]}\mathrm{diam}(\Omega_k)\big)  \E\bigg[\frac{M^2}{N^2}\bigg] + \tilde{C}_2 \E\bigg[2 - \frac{M}{N^2}\bigg] \bigg(1 -\sum_{k\in[m]}  \Big(  \frac{n_k}{n}\wedge \hat{\nu}_k \Big) \bigg)+ \tilde{C}_2\E\bigg[1 - \frac{M^2}{{N^2}}\bigg]\,  \bigg],
\end{align*}
where we used that $$\E_{\lambda}(L^2\mvert M) = {M}\Prob_{\lambda}( \mathbb{Z}^1\neq (0,\ldots,0)) + ({M}^2-{M}) \Prob_{\lambda}( \mathbb{Z}^1\neq (0,\ldots,0))^2.$$
Furthermore, the second last inequality follows as $(a^2-b^2) = (a-b)(a+b) \leq 2\max\{a,b\}(a-b)$ while the last inequality uses that $N$ and $M$ are independent of $\lambda$. 

Noting that $1 = \sum_{k\in[m]}\frac{n_k}{n}$, we obtain
$$\E\bigg[1- \sum_{k\in[m]}\frac{n_k}{n}\wedge \hat{\nu}_k\bigg] = \E\bigg[\sum_{k\in[m]}\frac{n_k}{n}- \big(\frac{n_k}{n}\wedge \hat{\nu}_k\big)\bigg] =\E\bigg[ \sum_{k\in[m]}\big(\frac{n_k}{n}- \hat{\nu}_k\big)_+\bigg]\leq \E\bigg[\sum_{k\in[m]}\abs{\frac{n_k}{n}- \hat{\nu}_k}\bigg].$$
Now recall, that $\hnu$ was chosen as the probability measure closest to $\nu$ with respect to the total variation metric. In particular, we have
\begin{align*}
\E\bigg[\sum_{k\in[m]}\abs{\frac{n_k}{n}- \hat{\nu}_k} \bigg]
&\leq \E\bigg[\sum_{k\in[m]}\abs{\frac{n_k }{n}- \frac{n_k + \lambda_k}{n}}   + \sum_{k\in[m]}\abs{\frac{n_k + \lambda_k}{n}- \hat{\nu}_k}  \bigg]  \\
&\leq 2 \sum_{k\in[m]}\E\bigg[\abs{\frac{n_k }{n}- \frac{n_k + \lambda_k}{n}}\bigg] \\
&= \frac{2}{n} \sum_{k\in[m]}\E[\abs{\lambda_k}] = 2\frac{m}{n} \E(\abs{\Lambda}).
\end{align*}

\textit{Step 2: Calculation of Poisson Expectations}\\
It remains to calculate the expectations in the upper bound. Recall that we coupled $N\sim \text{Poi}(a)$ and $M\sim \text{Poi}(b)$ such that $N {= } M + L$ for $L\sim \text{Poi}(c)$ with $c=a-b$. In particular, $N$ and $M$ are independent of $\lambda$.

For $c>0$ we obtain
\begin{align*}
    \E\bigg[\frac{M}{M+L}\bigg] &= \sum_{l=0}^{\infty}e^{-c}\frac{c^l}{l!}\, \sum_{m=1}^{\infty} e^{-b} \frac{b^m}{m!} \frac{m}{m+l}\\
    &= \sum_{l=0}^{\infty}e^{-c}\frac{c^l}{l!}\, \sum_{m=1}^{\infty} e^{-b} \frac{b^m}{(m-1)!} \frac{1}{m+l}\\
    %&= \sum_{l=0}^{\infty}e^{-c}\frac{c^l}{l!}\, b\, \sum_{m=0}^{\infty} e^{-b} \frac{b^m}{m!} \frac{1}{m+1+l}\\
    &= b\, \sum_{l=0}^{\infty}e^{-c}\frac{c^l}{l!}\,  \E\bigg[\frac{1}{M+1+l}\bigg]\\
    %&= \frac{b}{c}\, \sum_{l=0}^{\infty}e^{-c}\frac{c^{l+1}}{(l+1)!}\,  \E\bigg[\frac{l+1}{M+l+1}\bigg]\\
    &=\frac{b}{c}\, \sum_{l=1}^{\infty}e^{-c}\frac{c^{l}}{l!}\,  \E\bigg[\frac{l}{M+l}\bigg]\\
    &=\frac{b}{c}\E\bigg[\frac{L}{M+L}\bigg] = \frac{b}{c}\E\bigg[1 - \frac{M}{M+L}\bigg].\\
\end{align*}
This implies $\E\bigg[\frac{M}{M+L}\bigg] =\frac{1}{1+\frac{b}{c}} \1_{\{c>0\}} + \1_{\{c=0\}}=\frac{1}{1+\frac{b}{c}\1_{\{c>0\}}} $.
 By Jensen's inequality, this yields
$$\E\bigg[1 - \frac{M^2}{N^2}\bigg] \leq  1 - \E\bigg[\frac{M}{N}\bigg]^2 = 1-\E\bigg[\frac{M}{M+L}\bigg]^2 \leq \bigg(1 - \frac{1}{\big(1+\frac{b}{c}\big)^2}\bigg) \1_{\{c>0\}}.$$
Furthermore, as $M/(M+L)\leq 1$ we have
$$\E\bigg[\frac{M^2}{N^2}\bigg] = \E\bigg[\frac{M^2}{(M+L)^2}\bigg]\leq \E\bigg[\frac{M}{M+L}\bigg] \leq \frac{1}{1+\frac{b}{c}\1_{\{c>0\}}} .$$

Finally, the upper bound obtained in Step~4 yields
\begin{align*}
    &\E(d_{FGW}(\eta^{(\hat{\Eta},\hat{\Tau})},\eta^{(\hat{\Xi},\hat{\Sigma})}))  \\
    &\leq \tilde{C}_1\big(\max_{k\in[m]}\mathrm{diam}(\Omega_k)\big)  \E\bigg[\frac{M^2}{N^2}\bigg] + \tilde{C}_2  \E\bigg[2 - \frac{M}{N^2}\bigg] \E\bigg[1 -\sum_{k\in[m]}  \frac{n_k}{n}\wedge \hat{\nu}_k\bigg]\\
    &\hspace*{0.3cm}+ \tilde{C}_2 \E\bigg[1 - \frac{M^2}{{N^2}}\bigg],\\
    &\leq \tilde{C}_1\big(\max_{k\in[m]}\mathrm{diam}(\Omega_k)\big)  \frac{1}{1+\frac{b}{c}\1_{\{c>0\}}} + 4\tilde{C}_2 \frac{m}{n} \E(\abs{\Lambda})+ \tilde{C}_2 \bigg(1 - \frac{1}{\big(1+\frac{b}{c}\big)^2}\bigg) \1_{\{c>0\}}.
\end{align*}

\end{proof}
\newpage
% \subsection{Proof of Theorem~\ref{thm: result Stein}} \label{ssec: proof Stein}
\begin{tcolorbox}
\begin{reptheorem}{thm: result Stein}
Let $(P^{\mu^a_{\mcx}}\otimes Q^{\kappa})$ and $(P^{\mu^b_{\mcy}}\otimes Q^{\kappa})$ be distributions of the ``true'' and the private synthetic random graph of expected size $a>0$ and $b>0$, respectively, constructed as in Section~\ref{ssec: graph model} by using a probability measure $\hnu$ created by Algorithm~\ref{alg: PSMM} with random noise $\Lambda$.
Then 
    \begin{align*}
        &d_{\mcf}(P^{\mu^b_{\mcy}}\otimes Q^{\kappa} ,P^{\mu^a_{\mcx}}\otimes Q^{\kappa})\\
        &\leq  \bigg(1+\frac{1}{1+\frac{b}{c}\1_{\{c>0\}}}\bigg) (1-\alpha) \, \max_{k\in[m]}\mathrm{diam}(\Omega_k)+\tilde{C}_2\bigg(1 - \frac{1}{\big(1+\frac{b}{c}\big)^2}\bigg) \1_{\{c>0\}}\\
        &+2 c_V(a\wedge b)\frac{a\wedge b}{n}\Leb^d(\Omega) \E\abs{\Lambda} + c_E(a\wedge b)\,2L_{\kappa}\max_{k\in[m]}\mathrm{diam}(\Omega_k)^3 (a\wedge b)^2,\\
    \end{align*}
where 
\begin{align*}
    c_V(c) &:= \min \biggl\{ C_{\alpha}, % B^{*}
    \frac{1}{c} \,\bigl(1 +  ({1-e^{-{c}}}) \log^+{c} \bigr)\,C_{\alpha} \biggr\},\\
    c_E(c) &:= \min \biggl\{ 1, \frac{2-e^{-c}}{c} - \frac{1}{c^2}\Bigl(\frac{3}{2} - e^{-c}\Bigr) \biggr\} \; \alpha C
\end{align*}
for $C_{\alpha} := (1-\alpha)\mathrm{diam}(\Omega) + \alpha C$ with $\alpha\in [0,1]$ and $C>0$ parameters of the FGW metric and $L_{\kappa}$ the Lipschitz constant of the edge connection function~$\kappa$
\end{reptheorem}
\end{tcolorbox}

\begin{proof}
The strategy of this proof is as follows. We start by introducing so called intermediate graphs (of the same size) that are an auxiliary tool to handle the different support of $\Eta$ (having attributes in $\mcx$) and $\Xi$ (having attributes in $\mcy$). This is necessary as the general bounds in \citet[Theorem~4.7]{schuhmacher2024} are of total variation type and thus will not vanish in a direct comparison. Furthermore, we bound the distance of the intermediate graphs to their respective original graph.
We then compare the two intermediate graphs using a special case of \citet[Theorem~4.7]{schuhmacher2024}. Finally, we assemble all results to obtain to obtain the claimed upper bound. We write $\Prob_{\lambda}$ and $\E_{\lambda}$ to denote the conditional probability and expectation given $\lambda = (\lambda_i)_{i=1}^m$.

\textit{Step 1: Intermediate graphs}\\
Recall from Section~\ref{ssec: graph model}, that we have $N\sim \text{Poi}(a)$ %many 
vertices in $\Eta$ and $M\sim \text{Poi}(b)$ %many
vertices in $\Xi$;  assume w.l.o.g. that $a\geq b$. Then $N \overset{d}{=} M + K$ for $K\sim \text{Poi}(a-b)$. In particular, we can couple $N$ and $M$ such that $N\geq M$.

The intermediate graphs are obtained by resampling the (first) $M$ vertices uniformly in their corresponding partition set while keeping the edge probabilities of the original graph. We start with describing the first intermediate graph $(\Eta_I,\Tau_I)$ corresponding to the ``true'' graph $(\Eta,\Tau)$. Let the vertex process $\Eta_I$ be given by 
$$\Eta_I = \sum_{i=1}^m\sum_{j=1}^{N^I_i} \delta_{(X^I_{ij},U^I_{ij})}$$ where $X^I_{ij}\sim\mathcal{U}(\Omega_i)$, $U^I_{ij}\sim\mcu([0,1])$ and $N^I_i\sim\text{Poi}(\frac{n_i}{n}\,b)$ are independent of other random variables. %and $L$ the index of the maximal entry of $\Z = (Z_1,\ldots,Z_m)\sim \text{Multin}(1,(\frac{n_1}{n},\ldots,\frac{n_m}{n}))$. 
Furthermore, define the edge process $\Tau_I$ by
$$\Tau_I = \sum_{i,i'=1}^m\sum_{j=1}^{N^I_i}\sum_{\substack{j'=1}}^{N^I_{i'}}\tilde{K}_{iji'j'}^{\Tau_I} \delta_{\{U^I_{ij},U^I_{i'j'}\}}$$ for $\tilde{K}_{iji'j'}^{\Tau_I} \overset{i.i.d.\,}{\sim} \text{Ber}(\kappa(\tilde{X}_{ij},\tilde{X}_{i'j'}))$, where $\tilde{X}_{kl}\sim \mcu(\mcx \cap \Omega_k)$ for $k\in[m]$ and $l\in[N^I_k]$. We denote the distribution of the first intermediate graph $(\Eta_I,\Tau_I)$ by $P_I^{\mu^a_{\mcx}}\otimes Q_I^{\kappa_{\Tau}}$.

We construct a coupling $((\hat{\Eta},\hat{\Tau}),(\hat{\Eta}_I,\hat{\Tau}_I))$ of the intermediate graph $(\Eta_I,\Tau_I)$ and the ``true'' graph $(\Eta,\Tau)$ by setting $\hat{\Eta} = \Eta$ with $N = M + L$ for $L\sim\text{Poi}(a-b)$ and $\hat{\Tau} = \Tau$. Furthermore, we define $\hat{\Eta}_I := \sum_{i=1}^M \delta_{(\hat{X}_i^I,U_{ij})}$ for $\hat{X}_i^I\sim \mcu(\Omega_l)$ whenever $X_i\in\Omega_l$ and set $\hat{\Tau}_I = \sum_{\substack{i,j=1\\i\neq j}}^M K_{ij}^{\Tau} \delta_{\{\hat{U}_i,\hat{U}_j\}}$. In particular, the intermediate graph is obtained by uniformly resampling each attribute in the corresponding partition cell while keeping the edges fixed.

It remains to show that this defines a coupling, i.e. $(\hat{\Eta}_I,\hat{\Tau}_I) \overset{d}{=} (\Eta_I,\Tau_I)$. Let $\hat{N}_i := \hat{\Eta}_I(\Omega_i)$ be the number of vertices in $\hat{\Eta}_I$ that are in $\Omega_i$. Recall from Section~\ref{ssec: coupling approach} that the numbers of vertices $\N \mvert N :=(N_1,\ldots,N_m)$ of $\Eta$ in each partition set is multinominal distributed, i.e. $\N = \sum_{j=1}^N \N^j$ for $\N^j= (N^j_1,\ldots,N^j_m)\sim\text{Multin}(1,(\frac{n_1}{n},\ldots,\frac{n_m}{n}))$ independently. In particular, the construction of $\hat{\Eta}_I$ yields that for any $k_1,\ldots,k_m\in \Z$ with $k:=\sum_{i=1}^m k_i$ we have
\begin{align*}
    \Prob(\hat{N}_1 = k_1,\ldots \hat{N}_m = k_m) &= \Prob\bigg(\sum_{j=1}^M(N^j_1,\ldots,N^j_m)= (k_1,\ldots,k_m)\bigg) \\
    &= \sum_{l=0}^{\infty} \Prob(M = l) \Prob\bigg(\sum_{j=1}^M(N^j_1,\ldots,N^j_m)= (k_1,\ldots,k_m)\,\bigg\vert\, M=l\bigg) \\
    &= \Prob(M = k) \Prob\bigg(\sum_{j=1}^k(N^j_1,\ldots,N^j_m)= (k_1,\ldots,k_m)\bigg) \\
    &= e^{-b}\frac{b^k}{k!}\,\frac{k!}{k_1!\cdots k_m!} \prod_{i=1}^m \bigg(\frac{n_i}{n}\bigg)^{k_i}  \\
    %&= e^{-b \sum_{i=1}^m \frac{n_i}{n}}\frac{b^k}{k_1!\cdots k_m!} \prod_{i=1}^m \bigg(\frac{n_i}{n}\bigg)^{k_i}\\
    &= \prod_{i=1}^m e^{-\frac{n_i}{n}b}\frac{\big(\frac{n_i}{n}b\big)^{k_i}}{k_i!}\\
    &= \Prob(N^I_1 = k_1,\ldots N^I_m = k_m),
\end{align*}
where we used the independence of the $N_i^I$ for the last equation. This implies ${\Eta}_I \overset{d}{=} \hat{\Eta}_I$ as in both cases vertices were chosen uniformly from the partition cell. Furthermore, ${\Tau}_I \overset{d}{=} \hat{\Tau}_I$ as $\tilde{X}_{kl} \overset{d}{=} X_i\mvert (X_i\in \Omega_k)$. 

We aim to bound the expected distance of the intermediate graph and the ``true'' graph with respect to the FGW distance. Consider the coupling $((\hat{\Eta},\hat{\Tau}),(\hat{\Eta}_I,\hat{\Tau}_I))$ and note that $\hat{\Eta}_I$ is obtained from $\hat{\Eta}$ by resampling the first $M$ vertices in $\hat{\Eta}$ uniformly in its corresponding partition cell. In particular, $d(X_i,\hat{X}_i^I) \leq \max_{k\in[m]}\mathrm{diam}(\Omega_k)$ for all $i\in [M]$. Furthermore, we constructed $\hat{\Tau}_I$ in such a way that the edges are the same as in $\hat{\Tau}$. In particular, we obtain $d_{\hat{\Tau}}(U_i,U_j) - d_{\hat{\Tau}_I}(U_i,U_j) = 0$ for all $i,j\in[M]$.

We derive an upper bound of the FGW distance by restricting to transport plans that move mass $1/N$ from vertex $X_{i}$ (having mass $1/N$) to vertex $\hat{X}_i^I$ (having mass $1/M$) for all $i\in[M]$ and distribute the remaining mass $(1-M/N)$ according to some subcoupling of the remaining vertices. Note that this is analogous to the approach in Step~3 in the proof of Theorem~\ref{thm: result coupling}. Then for $\eta^{(\hat{\Eta},\hat{\Tau})}$ and $\eta^{(\hat{\Eta}_I,\hat{\Tau}_I)}$ the random measures that capture the (attributed) graph information of $(\hat{\Eta},\hat{\Tau})$ and $(\hat{\Eta}_I,\hat{\Tau}_I)$ we obtain
\begin{align*}
     &\E[\E(d_{FGW}(\eta^{(\hat{\Eta},\hat{\Tau})},\eta^{(\hat{\Eta}_I,\hat{\Tau}_I)})\mvert N,M)]\\
    %&\leq \E\bigg[\frac{1}{N^2}\bigg(\sum_{i=1}^M \E\big((1-\alpha)d(X_i,\hat{X}_i^I) \big)\bigg)^2 + \bigg(1-\frac{M^2}{N^2}\bigg)\tilde{C}_2\bigg] \\
    &\leq \E\bigg[\frac{M^2}{N^2}\bigg](1-\alpha)(\max_{k\in[m]}\mathrm{diam}(\Omega_k)) + \tilde{C}_2\E\bigg[1-\frac{M^2}{N^2}\bigg]\\
    & \leq \frac{1}{1+\frac{b}{c}\1_{\{c>0\}}}(1-\alpha)(\max_{k\in[m]}\mathrm{diam}(\Omega_k)) + \tilde{C}_2\bigg(1 - \frac{1}{\big(1+\frac{b}{c}\big)^2}\bigg) \1_{\{c>0\}},
\end{align*} 
for $\tilde{C}_2 =  (1-\alpha)\mathrm{diam}(\Omega)+\alpha \min\{C,2\,C\,L_{\kappa} \mathrm{diam}(\Omega)\} $ and $c=b-a$. We refer to Step~1 and Step~2 in the proof of Theorem~\ref{thm: result coupling} for details.

We define the second intermediate graph $(\Xi_I,\Sigma_I)$ in an analogous way. That is, we set 
$$\Xi_I = \sum_{i=1}^m\sum_{j=1}^{M_i} \delta_{(Y^I_{ij},V^I_{ij})},$$ where $V^I_{ij}\sim\mcu([0,1])$ and $Y^I_{ij}\sim\mathcal{U}(\Omega_i)$ for $j\in [M_i]$ with $M_i\sim\text{Poi}(\hat{\nu}_i b)$. %and $L$ the index of the maximal entry of $\Z = (Z_1,\ldots,Z_m)\sim \text{Multin}(1,(\hat{\nu}_1,\ldots,\hat{\nu}_m))$.
Furthermore, we define the edge process $\Sigma_I$ by
$$\Sigma_I = \sum_{i,i'=1}^m \sum_{j=1}^{M_i}\sum_{\substack{j'=1}}^{M_{i'}} \tilde{K}_{ij}^{\Sigma_I} \delta_{\{V^I_i,V^I_j\}}$$ for $\tilde{K}_{iji'j'}^{\Sigma_I} \overset{i.i.d.\,}{\sim} \text{Ber}(\kappa(y_i,y_{i'}))$. We denote the distribution of the second intermediate graph $(\Xi_I,\Sigma_I)$ by $P_I^{\mu^b_{\mcy}}\otimes Q_I^{\kappa_{\Sigma}}$. 
Analogously to the study of the first intermediate graph we can couple $(\Xi_I,\Sigma_I)$ and $(\Xi,\Sigma)$ such that $\abs{\Xi} = M = \sum_{i=1}^m M_i = \abs{\Xi_I}$. In particular, we compare two graphs of the same size and the bound derived for the first intermediate graph simplifies to
\begin{align*}
     \E(d_{FGW}(\eta^{(\Xi,\Sigma)},\eta^{(\Xi_I,\Sigma_I)})) 
    \leq (1-\alpha)(\max_{k\in[m]}\mathrm{diam}(\Omega_k)) .
\end{align*}

\textit{Step 2: Comparison of the intermediate graphs}\\
Consider the intermediate graphs and note that both vertex processes $\Eta_I$ and $\Xi_I$ are Poisson point processes,  with intensity functions $\lambda_{\Eta_I}(x) = \frac{n_i}{n}b  $ and $\lambda_{\Xi_I}(x) = \hat{\nu}_ib$ for $x\in\Omega_i$, respectively. Furthermore, the edges are conditionally (on the vertex attributes) independent. In particular, both intermediate graphs are so called generalized random geometric graphs, see \citet[Section 2.2]{schuhmacher2024}. Thus, we can apply \citet[Theorem 4.9]{schuhmacher2024} in the special case, where the third term vanishes, see \citet[Corollary 3.2] {schuhmacher2024}. This yields 
\begin{align} \label{eq: general approx result}
   & d_{\mcf}(P_I^{\mu^b_{\mcy}}\otimes Q_I^{\kappa_{\Sigma}} ,P_I^{\mu^a_{\mcx}}\otimes Q_I^{\kappa_{\Tau}}) \\&\leq c_V(b)\E \biggl(\int_{[0,1]}\int_{\Omega}\bigl|\lambda_{\Xi_I}(x) -\lambda_{\Eta_I}(x)\bigr|\, \alpha(\diff x) \diff u \biggr) \notag \\
    &\hspace*{4mm} {} + c_E(b)\,\E\biggl(\int_{[0,1]}\int_{\Omega} \sum_{(y,v)\in \Xi_I} \abs{\kappa_{\Tau}((x,u),(y,v)) - \kappa_{\Sigma}(x,y)} \,\lambda_{\Eta_I}(x) \,\alpha(\diff x) \diff u\biggr) \notag,%\\
    %&\hspace*{4mm} {} + \E \biggl( \int_{\Omega}\int_{\mfn_2} \triangle_Vh_f(\Xi_I,\sigma)\,\bigabs{\tilde{Q}_1(\Xi,x;\diff \sigma) - Q(\Xi_I;\diff \sigma)}\,\lambda_{\Xi_I}(\diff x \,\vert \Xi_I) \biggr),
\end{align}
where $\kappa_{\Tau}((x,u),(y,v)) = \kappa_{\Tau}(x,y)$ and $\kappa_{\Sigma}((x,u),(y,v)) = \kappa_{\Sigma}(x,y)$ denote the probabilities of an edge~${u,v}$ to be in $\Tau$ and $\Sigma$, respectively. Furthermore, 
\begin{align*}
    c_V(b) &= \min \biggl\{ C_{\alpha}, % B^{*}
    \frac{1}{b} \,\bigl(1 +  ({1-e^{-{b}}}) \log^+{b} \bigr)\,C_{\alpha} \biggr\},\\
    c_E(b) &= \min \biggl\{ 1, \frac{2-e^{-b}}{b} - \frac{1}{b^2}\Bigl(\frac{3}{2} - e^{-b}\Bigr) \biggr\} \; \alpha C
%&\overline{\triangle}_E\,h_f(\xi,\sigma)\leqc_E(\lambda) = \frac{1}{\Lambda} \,\bigl(1 +  ({1-e^{-\Lambda}}) \log^+{\Lambda} \bigr)\,C_E.
\end{align*}
for $C_{\alpha} := (1-\alpha)\mathrm{diam}(\Omega) + \alpha C$ an upper bound of $d_{FGW}$.  

Note that in contrast to \citet{schuhmacher2024} we chose the FGW metric $d_{FGW}$ instead of the GOSPA metric $d_{\G}$ as a underlying metric for the Wasserstein metric. However, as $d_{FGW}\leq d_{\G}$ (assuming that in both cases the vertex part is bounded by $(1-\alpha)\mathrm{diam}(\Omega)$ while edge part is bounded by $\alpha C$; compare \cite{vayer2020} and \cite{schuhmacher2023}), we obtain the same type of bounds.% for the Stein factors.

Applying the Mecke formula, see for example \citet[Corollary 31.11]{kallenberg1997}, %\gr{give citation for the Mecke formula?} 
and using that $\lambda_{\Eta_I}(x) =\frac{n_i}{n}b$ and $\lambda_{\Xi_I}(x) = \hat{\nu}_ib$ for $x\in\Omega_i$, we obtain for  $\tilde{X}_i\sim \mcu(\mcx\cap \Omega_i)$
\begin{align*}
    &d_{\mcf}(P_I^{\mu^b_{\mcy}}\otimes Q_I^{\kappa_{\Sigma}} ,P_I^{\mu^a_{\mcx}}\otimes Q_I^{\kappa_{\Tau}})\\
    &\leq
    c_V(b)\,\E \biggl(\int_{\Omega}\bigl|\lambda_{\Xi_I}(x) -\lambda_{\Eta_I}(x)\bigr|\, \alpha(\diff x) \biggr) \notag \\
    &\hspace*{4mm} {} + c_E(b)\,\int_{\Omega} \int_{\Omega} \E \bigg[\E_{\lambda}\bigl\vert\kappa_{\Tau}(x,y) - \kappa_{\Sigma}(x,y)\bigr\vert \,\lambda_{\Eta_I}(x)\,\lambda_{\Xi_I}(y)\bigg] \,\alpha(\diff x)\,\alpha(\diff y)  \notag\\
    &= c_V(b)\,\E \biggl(\sum_{i=1}^m\int_{\Omega_i}\biggl|\frac{n_i}{n} -\hat{\nu}_i\biggr| \, b\, \alpha(\diff x) \biggr) \notag \\
    &\hspace*{4mm} {} + c_E(b)\,\sum_{i,j=1}^m\int_{\Omega_i} \int_{\Omega_j} \E_{\lambda}\abs{\kappa(\tilde{X}_{i},\tilde{X}_j) - \kappa(y_i,y_j)} \,\frac{n_i}{n}\,\E[\hat{\nu}_j]\, b^2\,\alpha(\diff x)\alpha(\diff y) \notag\\
    &= c_V(n) \sum_{i=1}^m\Leb^d(\Omega_i)\E\biggl|\frac{n_i}{n} -\hat{\nu}_i\biggr| \, b  \\
    &\hspace*{4mm}+ c_E(b)\, \sum_{i,j=1}^m\Leb^d(\Omega_i)\Leb^d(\Omega_j)\E_{\lambda}\abs{\kappa(\tilde{X}_{i},\tilde{X}_j) - \kappa(y_i,y_j)} \,\frac{n_i}{n}\,\E[\hat{\nu}_j]\, b^2\\
    &\leq c_V(b) \sum_{i=1}^m\Leb^d(\Omega_i)\E\biggl|\frac{n_i}{n} -\hat{\nu}_i\biggr| \, b \\
    &\hspace*{4mm}+ c_E(b)\, \sum_{i,j=1}^m\Leb^d(\Omega_i)\Leb^d(\Omega_j)\,2L_{\kappa}\max_{k\in[m]}\mathrm{diam}(\Omega_k) \,\frac{n_i}{n}\,\E[\hat{\nu}_j]\, b^2,
\end{align*}
where the last inequality follows by the Lipschitz property of $\kappa$, see Inequality~\eqref{eq: lipschitz argument kappa}.

\textit{Step 3: Derivation of the upper bound}\\
Finally, we use the results of Step~$1$ and Step~$2$ to obtain the claimed upper bound. An application of the triangle inequality yields 
\begin{align*}
    &d_{\mcf}(P^{\mu^b_{\mcy}}\otimes Q^{\kappa} ,P^{\mu^a_{\mcx}}\otimes Q^{\kappa})\\
    &\leq d_{\mcf}(P^{\mu^b_{\mcy}}\otimes Q^{\kappa} ,P_I^{\mu^b_{\mcy}}\otimes Q_I^{\kappa_{\Sigma}}) 
    + d_{\mcf}(P_I^{\mu^b_{\mcy}}\otimes Q_I^{\kappa_{\Sigma}} ,P_I^{\mu^a_{\mcx}}\otimes Q_I^{\kappa_{\Tau}}) 
    + d_{\mcf}(P_I^{\mu^a_{\mcx}}\otimes Q_I^{\kappa_{\Tau}} ,P^{\mu^a_{\mcx}}\otimes Q_I^{\kappa_{\Tau}})\\
    % &=  \sup\limits_{f\in\mcf} \big\vert \E(f(\Xi,\Sigma)) - \E(f(\Xi_I,\Sigma_I))\big\vert 
    % + d_{\mcf}(P_I^{\mu^b_{\mcy}}\otimes Q_I^{\kappa_{\Sigma}} ,P_I^{\mu^a_{\mcx}}\otimes Q_I^{\kappa_{\Tau}}) 
    % +\sup\limits_{f\in\mcf} \big\vert \E(f(\Eta_I,\Tau_I)) - \E(f(\Eta,\Tau))\big\vert\\
    &\leq \E(d_{FGW}((\Xi,\Sigma),(\Xi_I,\Sigma_I)))
    + d_{\mcf}(P_I^{\mu^b_{\mcy}}\otimes Q_I^{\kappa_{\Sigma}} ,P_I^{\mu^a_{\mcx}}\otimes Q_I^{\kappa_{\Tau}}) 
    +\E(d_{FGW}((\Eta,\Tau),(\Eta_I,\Tau_I)))\\
    &\leq \bigg(1+\frac{1}{1+\frac{b}{c}\1_{\{c>0\}}}\bigg) (1-\alpha) \, \max_{k\in[m]}\mathrm{diam}(\Omega_k)+\tilde{C}_2\bigg(1 - \frac{1}{\big(1+\frac{b}{c}\big)^2}\bigg) \1_{\{c>0\}}\\
    &+ c_V(b) \sum_{i=1}^m\Leb^d(\Omega_i)\E\biggl|\frac{n_i}{n} -\hat{\nu}_i\biggr| \, b + c_E(b)\, \sum_{i,j=1}^m\Leb^d(\Omega_i)\Leb^d(\Omega_j)\,2L_{\kappa}\max_{k\in[m]}\mathrm{diam}(\Omega_k) \,\frac{n_i}{n}\,\E[\hat{\nu}_j]\, b^2\\
    &\leq \bigg(1+\frac{1}{1+\frac{b}{c}\1_{\{c>0\}}}\bigg) (1-\alpha) \, \max_{k\in[m]}\mathrm{diam}(\Omega_k)+\tilde{C}_2\bigg(1 - \frac{1}{\big(1+\frac{b}{c}\big)^2}\bigg) \1_{\{c>0\}}\\
    &+ 2 c_V(b)\frac{b}{n}\Leb^d(\Omega) \E\abs{\Lambda}  + c_E(b)\,  2L_{\kappa}\max_{k\in[m]}\mathrm{diam}(\Omega_k)^3\, b^2,
\end{align*}
where the last inequality follows as analogous to the proof of Theorem~\ref{thm: result coupling}; we have
$$\sum_{i=1}^m\Leb^d(\Omega_i)\E\bigl|\frac{n_i}{n} -\hat{\nu}_i\bigr| \leq \frac{2}{n}\sum_{i=1}^m\Leb^d(\Omega_i)\E\abs{\Lambda} = \frac{2}{n}\Leb^d(\Omega)\E\abs{\Lambda}.$$ 
\end{proof}

\subsection{Further tables} 

Here we give further tables on explicit bounds, see Section~\ref{ssec: choice of parameters}. The results show a similar picture as in Table \ref{tab: explicit bounds}; Corollary~\ref{cor: special case coupling bounds} and
Corollary~\ref{cor: special case Stein} give bounds of the same order. Furthermore, the bounds decrease with increasing $n$. A small $\alpha$ is favorable for both, Corollary~\ref{cor: special case Stein} and Corollary~\ref{cor: special case coupling bounds}. Overall the results illustrate that theoretic results obtained in Section~\ref{ssec: choice of parameters}.

\begin{table}[ht!]
    \centering
    \begin{filecontents*}{Tables/tableBoundCompOptfnAlpha14.csv}
"";"as.character.eps.1..";"X100";"X100.1";"X1000";"X1000.1";"X10000";"X10000.1"
"X.";"eps";"n = 100";"n = 100";"n = 1000";"n = 1000";"n = 10000";"n = 10000"
"X1";"1";"0.697";"0.574";"0.324";"0.254";"0.151";"0.117"
"X2";"0.1";"1.487";"1.378";"0.697";"0.574";"0.324";"0.254"
"X3";"0.01";"2.884";"3.367";"1.487";"1.378";"0.697";"0.574"
\end{filecontents*}
\pgfplotstabletypeset[
        col sep=semicolon,
        string type,
        column type=c,
        string replace*={"}{},
        header=false,
        every head row/.style={ 
            output empty row,
            before row={%
                \hline
                \multicolumn{1}{|c}{}&
                \multicolumn{1}{c|}{}&
                \multicolumn{6}{c|}{$n$}\\[0.5ex]
                \cline{3-8}
                \multicolumn{1}{|c}{}&
                \multicolumn{1}{c|}{}&
                \multicolumn{2}{c|}{100}&
                \multicolumn{2}{c|}{1000}&
                \multicolumn{2}{c|}{10.000}\\[0.5ex]
                \cline{3-8}
                \multicolumn{1}{|c}{}&
                \multicolumn{1}{c|}{}&
                \multicolumn{1}{c|}{Cor~\ref{cor: special case coupling bounds}}&
                \multicolumn{1}{c|}{Cor~\ref{cor: special case Stein}}&
                \multicolumn{1}{c|}{Cor~\ref{cor: special case coupling bounds}}&
                \multicolumn{1}{c|}{Cor~\ref{cor: special case Stein}}&
                \multicolumn{1}{c|}{Cor~\ref{cor: special case coupling bounds}}&
                \multicolumn{1}{c|}{Cor~\ref{cor: special case Stein}}\\
                \hline
            }
        },
        create on use/newCol/.style={
            create col/set list={" "," "," ",$\epsilon$, " "}
        },
        skip rows between index={0}{2},
        columns/newCol/.style={
            string type,
            column type/.add={|}{}
        },
        columns={newCol, [index]1, [index]2, [index]3, [index]4, [index]5, [index]6, [index]7},
        every last row/.style={after row=\hline},
        every odd column/.style={column type/.add={}{|}},
        every even column/.style={column type/.add={}{|}},
        every row/.style={before row={\rule{0pt}{2.5ex}}}
    ]{Tables/tableBoundCompOptfnAlpha14.csv}
    \caption{Explicit bounds obtained in Corollary~\ref{cor: special case coupling bounds} and Corollary~\ref{cor: special case Stein} assuming discrete Laplacian noise and $\Omega = [0,1]^2$, $\alpha = 1/4$ as well as $C = L_{\kappa} = 1$. Furthermore, $m = \lceil f_n(\epsilon)n\rceil$ with $f_n$ optimal and $a$ optimal w.r.t. Section~\ref{ssec: choice of parameters}.}
\end{table}

\begin{table}[ht!]
    \centering
    \begin{filecontents*}{Tables/tableBoundCompOptfnAlpha34.csv}
"";"as.character.eps.1..";"X100";"X100.1";"X1000";"X1000.1";"X10000";"X10000.1"
"X.";"eps";"n = 100";"n = 100";"n = 1000";"n = 1000";"n = 10000";"n = 10000"
"X1";"1";"0.804";"0.787";"0.374";"0.354";"0.174";"0.163"
"X2";"0.1";"1.711";"1.825";"0.804";"0.787";"0.374";"0.354"
"X3";"0.01";"3.237";"4.074";"1.711";"1.825";"0.804";"0.787"
\end{filecontents*}
\pgfplotstabletypeset[
        col sep=semicolon,
        string type,
        column type=c,
        string replace*={"}{},
        header=false,
        every head row/.style={ 
            output empty row,
            before row={%
                \hline
                \multicolumn{1}{|c}{}&
                \multicolumn{1}{c|}{}&
                \multicolumn{6}{c|}{$n$}\\[0.5ex]
                \cline{3-8}
                \multicolumn{1}{|c}{}&
                \multicolumn{1}{c|}{}&
                \multicolumn{2}{c|}{100}&
                \multicolumn{2}{c|}{1000}&
                \multicolumn{2}{c|}{10.000}\\[0.5ex]
                \cline{3-8}
                \multicolumn{1}{|c}{}&
                \multicolumn{1}{c|}{}&
                \multicolumn{1}{c|}{Cor~\ref{cor: special case coupling bounds}}&
                \multicolumn{1}{c|}{Cor~\ref{cor: special case Stein}}&
                \multicolumn{1}{c|}{Cor~\ref{cor: special case coupling bounds}}&
                \multicolumn{1}{c|}{Cor~\ref{cor: special case Stein}}&
                \multicolumn{1}{c|}{Cor~\ref{cor: special case coupling bounds}}&
                \multicolumn{1}{c|}{Cor~\ref{cor: special case Stein}}\\
                \hline
            }
        },
        create on use/newCol/.style={
            create col/set list={" "," "," ",$\epsilon$, " "}
        },
        skip rows between index={0}{2},
        columns/newCol/.style={
            string type,
            column type/.add={|}{}
        },
        columns={newCol, [index]1, [index]2, [index]3, [index]4, [index]5, [index]6, [index]7},
        every last row/.style={after row=\hline},
        every odd column/.style={column type/.add={}{|}},
        every even column/.style={column type/.add={}{|}},
        every row/.style={before row={\rule{0pt}{2.5ex}}}
    ]{Tables/tableBoundCompOptfnAlpha34.csv}
    \caption{Explicit bounds obtained in Corollary~\ref{cor: special case coupling bounds} and Corollary~\ref{cor: special case Stein} assuming discrete Laplacian noise and $\Omega = [0,1]^2$, $\alpha = 3/4$ as well as $C = L_{\kappa} = 1$. Furthermore, $m = \lceil f_n(\epsilon)n\rceil$ with $f_n$ optimal and $a$ optimal w.r.t. Section~\ref{ssec: choice of parameters}}
\end{table}
\end{document}